\documentclass[11pt]{article}
\usepackage[numbers,compress]{natbib}
\usepackage[margin=1in]{geometry}
\usepackage[utf8]{inputenc} 
\usepackage[T1]{fontenc}    
\usepackage{hyperref}       
\usepackage{url}            
\usepackage{booktabs}       
\usepackage{amsfonts}       
\usepackage{nicefrac}       
\usepackage{microtype}      
\usepackage{graphicx}       
\usepackage{scalerel}       

\usepackage{amsmath}
\usepackage{amssymb}
\usepackage{amsthm}
\usepackage{bm,bbm}

\usepackage[usenames,dvipsnames]{xcolor}
\usepackage{bm}
\usepackage{bbm}
\usepackage{dsfont}
\usepackage{amsbsy}
\usepackage{amssymb}
\usepackage{amsmath}%
\usepackage{amsfonts}%
\usepackage{amsthm}
\usepackage{breqn}
\usepackage{graphicx}
\usepackage{colonequals}
\usepackage{psfrag}
\usepackage{subfigure}
\usepackage{mathrsfs}
\usepackage[noadjust]{cite}
\usepackage{hyperref}
\usepackage{mdframed}
\usepackage{enumerate}
\usepackage{enumitem}
\usepackage{braket}
\usepackage{nicefrac}       
\usepackage{mathtools}

\allowdisplaybreaks

\theoremstyle{plain}
\newtheorem{theorem}{Theorem}[section]
\newtheorem{proposition}[theorem]{Proposition}

\theoremstyle{definition}

\theoremstyle{remark}

\usepackage{sidecap}

\makeatletter
\newtheorem*{rep@theorem}{\rep@title}
\newcommand{\newreptheorem}[2]{%
\newenvironment{rep#1}[1]{%
 \def\rep@title{#2 \ref{##1}}%
 \begin{rep@theorem}}%
 {\end{rep@theorem}}}
\makeatother
\newreptheorem{conj}{Conjecture}

\theoremstyle{definition}

\newtheorem{defn}{Definition}

\newtheorem{remark}{Remark}
\newtheorem{example}{Example}

\newcommand{\BR}{\mathbb{R}}
\newcommand{\BE}{\mathbb{E}}
\newcommand{\BP}{\mathbb{P}}

\newcommand{\bgamma}{\boldsymbol{\gamma}}
\newcommand{\bDelta}{\boldsymbol{\Delta}}
\newcommand{\bh}{\boldsymbol{h}}

\newcommand{\ens}{\textup{ens}}

\newcommand{\calR}{\mathcal{R}}
\newcommand{\calA}{\mathcal{A}}

\newcommand{\calT}{\mathcal{T}}

\newcommand{\calS}{\mathcal{S}}

\newcommand{\calH}{\mathcal{H}}

\newcommand{\bx}{\boldsymbol{x}}

\newcommand{\bX}{\boldsymbol{X}}

\renewcommand{\tilde}{\widetilde}

\newcommand{\Reals}{\mathbb{R}}

\newcommand{\defined}{\triangleq}

\newcommand{\blambda}{\boldsymbol{\lambda}}

\newcommand{\indicator}{\mathds{1}}

\newcommand{\calD}{\mathcal{D}}

\newcommand{\hatR}{\hat{\calR}(\calH,\calD,\epsilon)}
\newcommand{\hatRm}{\hat{\calR}_m}

\definecolor{light-gray}{gray}{.90}

\newmdenv[%
  backgroundcolor=SkyBlue!30, %
  linecolor=black,
  linewidth =1pt,%
  skipabove = 10pt,%
  skipbelow = 10pt
]{comment}

\newmdenv[%
  backgroundcolor=red, %
  linecolor=black,
  linewidth =1pt,%
  skipabove = 10pt,%
  skipbelow = 10pt
]{TODO}

\usepackage{microtype}
\definecolor{quotemark}{gray}{0.7}
\makeatletter
\def\fquote{%
    \@ifnextchar[{\fquote@i}{\fquote@i[]}
           }%
\def\fquote@i[#1]{%
    \def\tempa{#1}%
    \@ifnextchar[{\fquote@ii}{\fquote@ii[]}
                 }%
\def\fquote@ii[#1]{%
    \def\tempb{#1}%
    \@ifnextchar[{\fquote@iii}{\fquote@iii[]}
                      }%
\def\fquote@iii[#1]{%
    \def\tempc{#1}%
    \vspace{1em}%
    \noindent%
    \begin{list}{}{%
         \setlength{\leftmargin}{0.1\textwidth}%
         \setlength{\rightmargin}{0.1\textwidth}%
                  }%
         \item[]%
         \begin{picture}(0,0)%
         \put(-15,-5){\makebox(0,0){\scalebox{3}{\textcolor{quotemark}{``}}}}%
         \end{picture}%
         \begingroup\itshape}%
 \def\endfquote{%
 \endgroup\par%
 \makebox[0pt][l]{%
 \hspace{0.8\textwidth}%
 \begin{picture}(0,0)(0,0)%
 \put(15,15){\makebox(0,0){%
 \scalebox{3}{\color{quotemark}''}}}%
 \end{picture}}%
 \ifx\tempa\empty%
 \else%
    \ifx\tempc\empty%
       \hfill\rule{100pt}{0.5pt}\\\mbox{}\hfill\tempa,\ \emph{\tempb}%
   \else%
       \hfill\rule{100pt}{0.5pt}\\\mbox{}\hfill\tempa,\ \emph{\tempb},\ \tempc%
   \fi\fi\par%
   \vspace{0.5em}%
 \end{list}%
 }%
 \makeatother

\usepackage{algorithm}
\usepackage{algpseudocode}
\usepackage{xcolor}
\hypersetup{%
  colorlinks=true,
  urlcolor=blue,%
  linkcolor=blue,%
  citecolor=blue,%
  linkbordercolor=blue,
  pdfborderstyle={/S/U/W 1}
}
\usepackage[labelfont=bf]{caption}

\usepackage{subfiles}
\title{Arbitrariness Lies Beyond the Fairness-Accuracy Frontier }
\date{}
\author{
Carol Xuan Long${}^\ddagger$, Hsiang Hsu\thanks{equal contributions.} \thanks{JPMorgan Chase Bank, N.A., New York, NY, USA. Email: \texttt{hsiang.hsu@jpmchase.com}.}, Wael Alghamdi${}^*$${}^\ddagger$, Flavio P. Calmon\thanks{John A. Paulson School of Engineering and Applied Sciences, Harvard University, Boston, MA 02134. Emails: \texttt{carol\_long@g.harvard.edu, alghamdi@g.harvard.edu, flavio@seas.harvard.edu}.}
}

\begin{document}

\maketitle

\begin{abstract}
Machine learning tasks may admit multiple competing models that achieve similar performance yet produce conflicting outputs for individual samples---a phenomenon known as predictive multiplicity. We demonstrate that fairness interventions in machine learning optimized solely for group fairness and accuracy can exacerbate predictive multiplicity. Consequently, state-of-the-art fairness interventions can mask high predictive multiplicity behind favorable group fairness and accuracy metrics. We argue that a third axis of ``arbitrariness'' should be considered  when deploying models to aid decision-making in applications of individual-level impact.
To address this challenge, we propose an ensemble  algorithm applicable to any fairness intervention that provably ensures  more consistent predictions. 
\end{abstract}


\section{Introduction}

Non-arbitrariness is an important facet of non-discriminatory decision-making. Substantial arbitrariness exists in the training and selection of machine learning (ML) models. 
By simply varying hyperparameters of the training process (e.g., random seeds in model training), we can produce models with conflicting outputs on individual input samples \citep{breiman2001statistical,marx2020predictive, hsu2022rashomon, cooper2023variance}. 
The phenomenon where distinct models exhibit similar accuracy but inconsistent individual predictions is called \textit{predictive multiplicity} \citep{marx2020predictive}.  The arbitrary variation of outputs due to unjustified choices made during training can disparately impact individual samples, i.e., predictive multiplicity is not equally distributed across inputs of a model. When deployed in high-stakes domains (e.g., medicine, education, resume screening), the arbitrariness in the ML pipeline may target and cause systemic harm to specific individuals by excluding them from favorable outcomes  \citep{creel2022algorithmic,black2022model,watson2022predictive}. 

Popular fairness metrics in the ML literature do not explicitly capture non-arbitrariness. A widely recognized notion of non-discrimination is \emph{group fairness}. Group fairness is quantified in terms of, for example, statistical parity~\citep{dwork2015preserving}, equal opportunity, equalized odds~\citep{hardt2016equality}, and  variations such as multiaccuracy~\citep{kim2019multiaccuracy} and multicalibration~\citep{hebert2018multicalibration}. Broadly speaking, methods that control for group fairness aim to guarantee comparable performance of a model across population groups in the data. The pursuit of group fairness has led to hundreds of fairness interventions that seek to control for performance disparities while preserving accuracy \citep{hort2022bia}.

The central question we tackle in this paper is: Do models corrected for group fairness exhibit less arbitrariness in their outputs? We answer this question in the \emph{negative}. We demonstrate  that state-of-the-art fairness interventions may improve group fairness metrics at the expense of exacerbating arbitrariness. The harm is silent: the increase in arbitrariness is masked by favorable group fairness and accuracy metrics. Our results show that arbitrariness lies beyond the fairness-accuracy frontier: predictive multiplicity should be accounted for \emph{in addition} to usual group-fairness and accuracy metrics during model development.

\begin{figure}[!tb]
\centering
\includegraphics[width=1\textwidth]{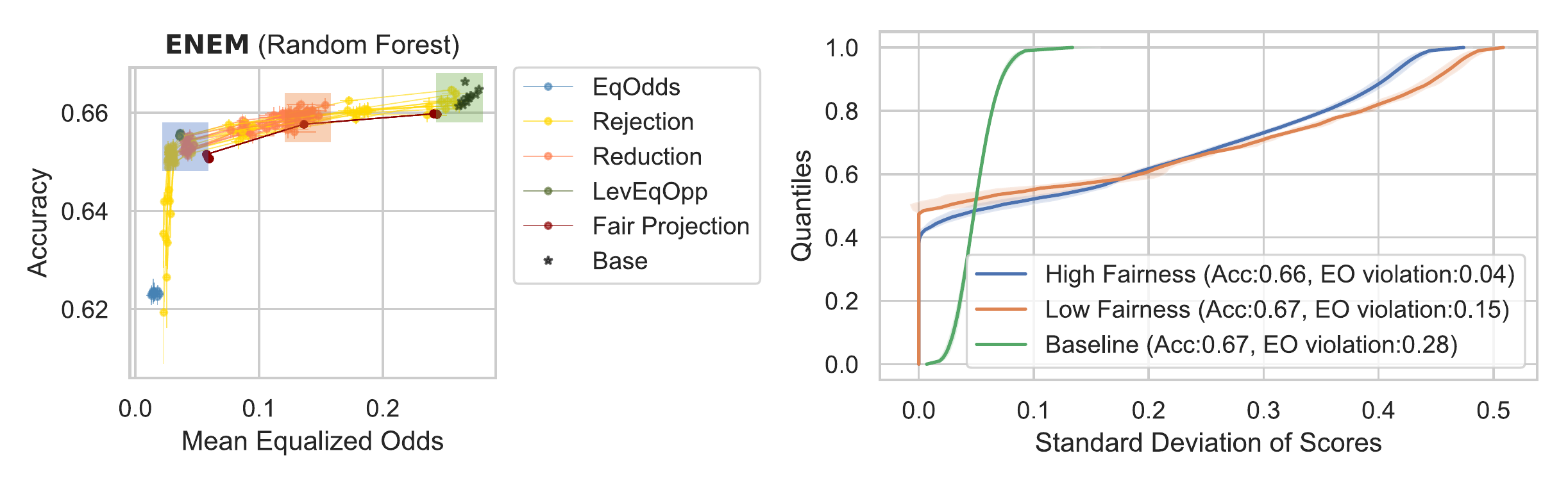}
\caption{\footnotesize
Accuracy-fairness frontier does not reveal arbitrariness in competing models. \textbf{Left}: Fairness-Accuracy frontier of baseline and fair models corrected by 5 fairness interventions; point clouds generated by different random seed choices.
\textbf{Right}: The cumulative distribution functions (CDF) of per-sample score std. across classifiers at different intervention levels (see Defn.~\ref{def:: score SD}). For each sample, std. is measured across competing scores produced by classifiers initialized with different random seeds. A \emph{wider} CDF indicates \emph{more} disparity of the impact of arbitrariness on different individuals.
}
\label{fig::Enem quantile}
\end{figure}

Figure~\ref{fig::Enem quantile} illustrates how fairness interventions can increase predictive multiplicity. Here, state-of-the-art fairness interventions are applied\footnote{See Section~\ref{sec:exp} for a detailed description of the experiment and dataset.} to a baseline random forest classifier to ensure group fairness (mean equalized odds \citep{hardt2016equality})  in a student performance prediction task (binary prediction). We produce multiple baseline classifiers by varying the random seed used to initialize the training algorithm. Each baseline classifier achieves comparable accuracy and fairness violation. They also mostly agree in their predictions: for each input sample, the standard deviation of output scores across classifiers is small (see Defn.~\ref{def:: score SD}). After applying a fairness intervention to each randomly initialized baseline classifier, we consistently reduce  group fairness violations at a small accuracy cost, as expected. However, predictive multiplicity changes significantly post-intervention: for roughly half of the students, predictions are consistent across seeds, whereas for 20\% of the students, predictions are comparable to a coin flip. For the latter group, the classifier output depends on the choice of a random seed instead of any specific input feature. The increase in predictive multiplicity is masked by the fairness-accuracy curve, does not impact all samples equally, and is consistent across datasets and learning tasks.

At first, the increase in predictive multiplicity may seem counter-intuitive: adding fairness constraints to a learning task should reduce the solution space, leading to less disagreement across similarly-performing classifiers relative to an unconstrained baseline. We demonstrate that, in general, this is not the case. For a given hypothesis class, the non-convex nature of group fairness constraints can in fact \emph{increase} the number of feasible classifiers at a given fairness and accuracy level. We show that this phenomenon occurs even in the simple case where the  hypothesis space is comprised of threshold classifiers over one-dimensional input features, and the optimal baseline classifier is unique. To address this challenge, we demonstrate -- both theoretically and through experiments -- that ensembling classifiers is an effective strategy to counteract this multiplicity increase.

The main contributions of this work include:
\begin{enumerate}
    \item We demonstrate that the usual ``fairness-accuracy'' curves can systematically mask an increase of predictive multiplicity.  Notably, applying state-of-the-art fairness interventions can incur higher arbitrariness in the ML pipeline.
   
    \item We show that multiplicity can be arbitrarily high even if group fairness and accuracy are controlled, provided that the models do not achieve 100\% test accuracy. Hence,  fairness interventions optimized solely for  fairness and accuracy cannot, in general, control predictive multiplicity. We also provide  examples on why fairness constraints may exacerbate arbitrariness.
    \item We propose an ensemble algorithm that reduces multiplicity while maintaining fairness and accuracy. We derive convergence rate results to show that the probability of models disagreeing drops exponentially as more models are added to the ensemble.
    \item We demonstrate the multiplicity phenomena and benchmark our ensemble method through comprehensive experiments using state-of-the-art fairness interventions across real-world datasets.
\end{enumerate}
Proofs and additional experiments are included in the Appendices.

\subsection{Related Work}
\paragraph{Multiplicity, its implications, and promises.} 
Recent works have investigated various factors that give rise to multiplicity. 
\citet{d2022underspecification} studied how under-specification presents challenges to the credibility of modern machine learning algorithms. 
\citet{creel2022algorithmic} thoroughly explored the notion of arbitrariness in machine learning and discuss how high-multiplicity predictions can lead to systematized discrimination in society through ``algorithmic leviathans.'' 
Multiplicity in prediction and classification can also have beneficial effects.
\citet{black2022model,semenova2019study}, and \citet{fisher2019all} view multiplicity of equally-performing models as an opportunity to optimize for additional criteria such as generalizability, interpretability, and fairness. 
\citet{coston2021characterizing} develop a framework to search over the models in the Rashomon set for a better operation point on the accuracy-fairness frontier. 
However, they do not discuss the potential predictive multiplicity cost of existing fairness interventions nor propose algorithms to reduce this cost.

The work most similar to ours is \citep{cooper2023variance}.
\citet{cooper2023variance} consider the problem of predictive multiplicity as a result of using different splits of the training data. Therein, they quantify predictive multiplicity by prediction variance, and they propose a  bagging strategy \citep{breiman1996bagging} to combine models.
Our work considers a different problem where predictive multiplicity is exacerbated by group-fairness interventions. 
Our work is also different from \citet{cooper2023variance} as we fix the dataset when training models and consider multiplicity due to randomness used during training. 
In this sense, our ensemble algorithm is actually a voting ensemble \citep{witten2002data} (see Section~\ref{sec:ensemble}); see also ensembling and reconciliation strategies proposed by \citet{black2021selective} and \citet{roth2022reconciling} that aim to create more consistent predictions among competing models. To the best of the authors' knowledge, we are the first to measure and report the arbitrariness cost of  fairness interventions.

\paragraph{Hidden costs of randomized algorithms.}
Recent works~\citep{ganesh2023impact,krco2023mitigating, kulynych2023arbitrary} examine the potential detrimental consequences of randomization in the ML pipeline.
In their empirical study, Ganesh et al.\citep{ganesh2023impact} observe that group fairness metrics exhibit high variance across models at different training epochs of Stochastic Gradient Descent (SGD). The authors point out that random data reshuffling in SGD makes empirical evaluation of fairness (on a test set) unreliable, and they attribute this phenomenon to the volatility of predictions in minority groups. Importantly, they do not incorporate fairness interventions in their experiments. In contrast, we apply fairness interventions to baseline models. Specifically, we examine the variance in predictions among models with similar fairness and accuracy performances. In addition to the observations made by Ganesh et al., our theoretically-grounded study reveals the different paths that lead to group-fairness, i.e., that arbitrariness can be an unwanted byproduct of imposing fairness constraints. \citet{krco2023mitigating} empirically study if fairness interventions reduce bias equally across groups, and examine whether affected groups overlap across different fairness interventions. 
In contrast, our work examines the \emph{multiplicity cost} of group fairness and its tension with individual-level prediction consistency, rather than the \emph{fairness cost} of  randomness in the ML pipeline. 
Another work on the hidden cost of randomized algorithms is given by \citet{kulynych2023arbitrary}, who report that  well-known differentially-private training mechanisms can exacerbate predictive  multiplicity.

\section{Problem Formulation and Related Works} \label{sec:setup}

We explain the setup and relevant definitions in this section.

\textbf{Prediction tasks.} 
We consider a binary  classification setting where the training examples consist of triplets $(\bX,S,Y)\stackrel{iid}{\sim}P_{\bX,S,Y} $; $\bX$ is an $\mathbb{R}^d$-valued feature vector, $S$ is a discrete random variable supported on $[K]\defined \{1,\cdots,K\}$  representing $K$ (potentially overlapping) group memberships, and $Y\in \{0,1\}$\footnote{We note that our setup can be readily extended to multi-class prediction.}.
We consider probabilistic classifiers in a hypothesis space $\calH$, where each $h\in \calH$ is a mapping $h:\mathbb{R}^d\to [0,1]$. 
Each classifier $h(\bx)$ aims to approximate $P_{Y|\bX=\bx}(1)$. 
The predicted labels $\hat{y}$ can be obtained by thresholding the scores, i.e., $\hat{y} = \indicator \{h(\bx) \ge 0.5\}$, where $\indicator\{\cdot\}$ is the indicator function. 
Finally, we denote $\bDelta_c$ the $c$-dimensional probability simplex. 

We fix a training dataset of $n$ i.i.d samples $\calD \defined \{(\bx_i, s_i, y_i) \}_{i = 1}^n$ drawn from $P_{\bX,S,Y}$, and a randomized training procedure $\calT$. 
Let $\calT(\calD)$ denote the set of all possible models $h:\BR^d\to [0,1]$ trained with procedure $\calT$ on the training set $\calD$. 
For example, different classifiers in $\calT(\calD)$ can be induced by using different random seeds at the beginning of the execution of procedure $\calT$ on $\calD$.

\textbf{Predictive multiplicity and the Rashomon set.}  
 
Given a loss function $\ell:[0,1]\times[0,1]\to \Reals^+$, a
hypothesis class $\calH$ and a  parameter $\epsilon>0$, we define the \emph{empirical Rashomon set}  of $m$ competing models via re-training as
\begin{equation}
   \hatRm(\calH, \calD, \epsilon) \triangleq \left\{ h_1, h_2, \cdots, h_m \sim \calT(\calD) \ : \ \frac{1}{|\calD|}\sum_{(\bx,s, y)\in \calD}\ell(h_i(\bx),y)\leq \epsilon, \forall i \in [m] \right\}.  
\end{equation}
Here, $\epsilon$ is an approximation parameter that determines the size of the set. The set can $\hatRm(\calH, \calD,\epsilon)$ be viewed as an approximation of the \emph{Rashomon set} of ``good'' models \citep{breiman2001statistical,watson2022predictive,marx2020predictive}. We omit the arguments of $\hatRm(\calH, \calD,\epsilon)$ when they are clear are implied from context.

There are various metrics to quantify predictive multiplicity across models in  $\hatRm(\calH, \epsilon)$  by either considering their output scores \citep{semenova2019study, watson2022predictive} or thresholded predictions \citep{marx2020predictive}.
We focus on ambiguity for evaluating the predictive multiplicity of thresholded predictions and standard deviation (std.) for output scores, i.e., when model outputs are in the interval $[0,1]$. 

\begin{defn}[Ambiguity] 
\label{def:: ambiguity}
Given a dataset with $n$ i.i.d samples $\calD \defined \{(\bx_i, s_i, y_i) \}_{i = 1}^n$ and an empirical Rashomon set $\hatRm $, the ambiguity of a dataset over $\hatRm$ is the proportion of points in the dataset that can be assigned a conflicting prediction by a competing classifier within $\hatRm$:
\begin{equation}
    \alpha_{\hatRm} \triangleq \frac{1}{n}\sum_{i = 1}^{n} \max_{g,g'\in \hatRm} \indicator [g(\bx_i)\neq g'(\bx_i)].
\end{equation}
\end{defn}

\begin{defn}[Quantiles of std. of Scores] 
\label{def:: score SD}
Given an empirical Rashomon set $\hatRm$ with $m$ competing models, each outputs a score $f_i(\bx)$ for sample $\bx\in\calD$, the empirical standard deviation (std.) of scores for $\bx$ is given by
\begin{equation}
    s_\bx = \sqrt{\frac{1}{m-1}\sum_{i=1}^m (f_i(\bx)-\bar{\mu})^2},
\end{equation}
where $\bar{\mu}$ denotes the empirical mean of the scores. 
To understand the std. of scores of the population, we consider the empirical quantile of the std. of the scores. Given a dataset with $n$ i.i.d samples $\calD \defined \{(\bx_i, s_i, y_i) \}_{i = 1}^n$, $t$ is the $q$-th quantile ($q\in[0,1]$) of the std. of the scores if:
\begin{equation}
    \hat{F}(t) = \frac{1}{|\calD|}\sum_{\bx \in \calD} \indicator [s_\bx \leq t] = q,
\end{equation}
where $\hat{F}(t)$ is the empirical cumulative distribution function. 
\end{defn}

To instantiate the definitions, consider a resume screening task where the algorithm decides whether or not to extend an interview opportunity. If we obtain t = 0.5 being the 90\% quantile of the std. of scores, i.e. $\hat{F}(t) = 0.9$, this implies that for 10\% of the individuals in the dataset, the predictions produced by the competing models are arbitrary and conflicting: regardless of the mean scores, with an std. of 0.5, there would exist models with scores falling above and below the one-half threshold, so the thresholded output can be both 0 (no interview) and 1 (offer interview).

\textbf{Group fairness.}  
We consider three  group fairness definitions for classification tasks---statistical parity (SP), equalized odds (EO), and overall accuracy equality (OAE) \citep{dwork2015preserving,hardt2016equality,pleiss2017fairness,chouldechova2017fair}. OAE and EO are defined below as they are used in the next sections, and we refer the reader to Appendix~\ref{sec:: discussion fairness} for the remaining definitions. 

\begin{defn}[Overall Accuracy Equality, OAE] \label{def::OAE}
    The predictor $\widehat{Y}$ satisfies overall accuracy equality (OAE) if its accuracy is independent of the group attribute: for all groups $s,s'\in [K]$,
    \begin{equation}
        \Pr(\widehat{Y} = Y \mid S = s) = \Pr(\widehat{Y} = Y \mid S = s').
    \end{equation} 
\end{defn}

For binary classification, SP boils down to requiring the average predictions to be equal across groups, while EO requires true positive rates (TPR) and false positive rates (FPR) to be calibrated. In this paper, we consider mean equalized odds (Mean EO): the 
average of absolute difference in FPR and TPR for unprivileged and privileged groups \citep{bellamy2019ai}:
\begin{equation} \label{eq::EO}
    \textsc{Mean EO} \defined \frac{1}{2}\left( |\textsc{TPR}_{S=0}-\textsc{TPR}_{S=1}| +|\textsc{FPR}_{S=0}-\textsc{FPR}_{S=1}| \right).
\end{equation}

To examine whether current fairness intervention methods lead to an exacerbation of multiplicity, we survey state-of-the-art intervention methods, including Reductions~\citep{agarwal2018reductions}, Fair Projection~\citep{alghamdi2022beyond}, Reject Options~\citep{kamiran2012decision}, and EqOdds~\citep{hardt2016equality}. We offer a brief discussion of their mechanism in Appendix \ref{sec:: discussion fairness}.


\section{Orthogonality of Fairness and Arbitrariness}
We discuss next why arbitrariness is a third axis not captured by fairness and accuracy. Models with similar fairness and accuracy metrics can differ significantly in predictions. 
When predictive multiplicity is high, choosing any of the competing models for deployment without justification leads to arbitrariness in the ML pipeline. 

\begin{figure}[!tb]
  \centering
  \includegraphics[width=0.65\textwidth]{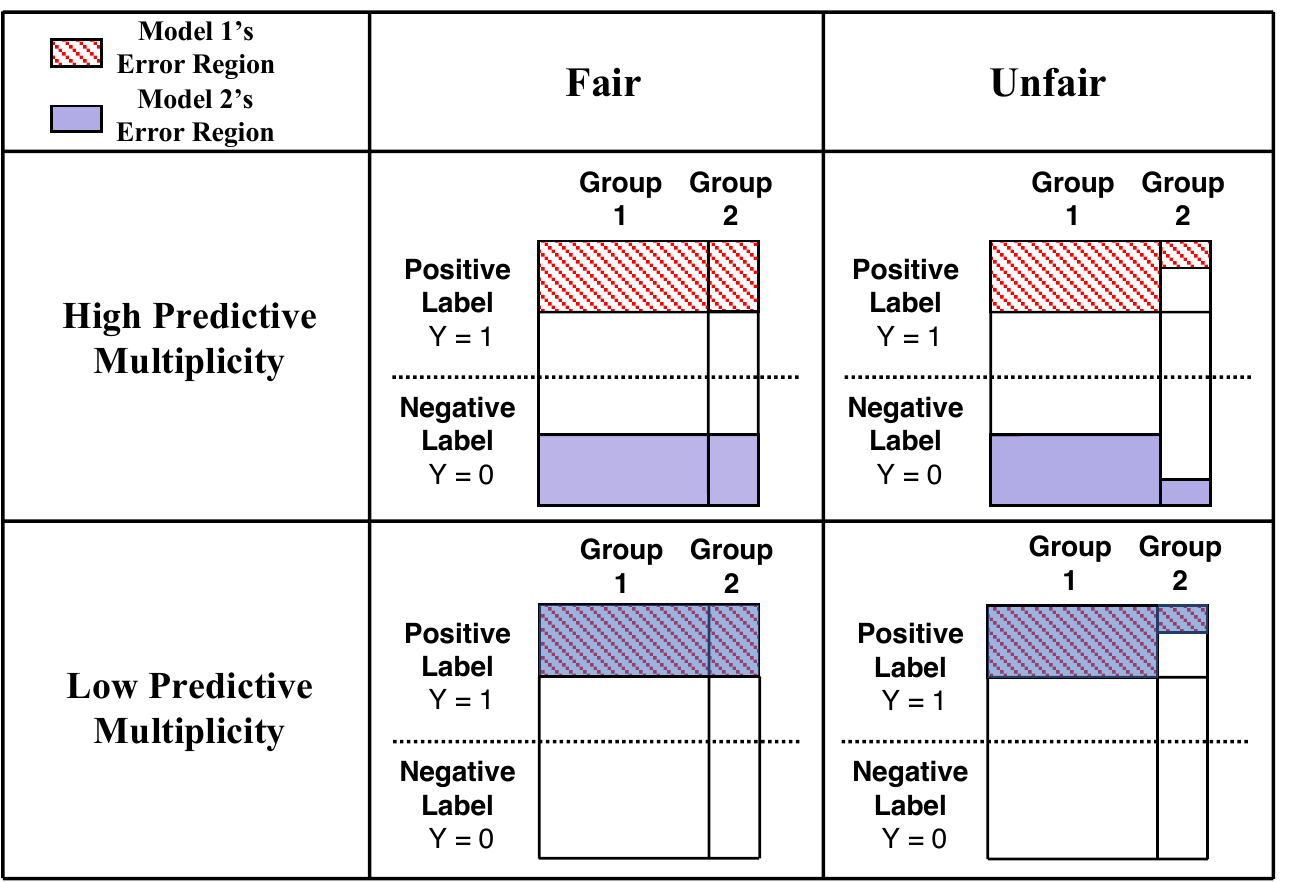}
  \caption{\footnotesize
  Illustration on two models being fair/unfair and exhibit high/low predictive multiplicity through the models' error regions in each of the 4 cases. The metrics for fairness and predictive multiplicity are Overall Accuracy Equality (Definition \ref{def::OAE}) and ambiguity (Definition \ref{def:: ambiguity}) respectively.}
\label{fig::diagram}
\end{figure}

\begin{example}[Ambiguity $\neq$ OAE]
 Overall Accuracy Equality (OAE, Definition \ref{def::OAE}) does not capture the ambiguity of model outputs (Definition  \ref{def:: ambiguity}). Consider two hypothetical models that are fair/unfair and exhibit high/low predictive multiplicity in Figure \ref{fig::diagram}. Here, in each panel, the rectangle represents the input feature space and the shaded regions represent the error region of each model.

In the top left panel, both Model 1 and 2 have equal accuracy for both groups, since the proportion of the error regions (red stripes and pink shade) for both groups are the same. Hence, both models are considered group-fair in terms of OAE. However, the error regions of the two models are disjoint. Since ambiguity is measured by the percentage of the samples that receive conflicting predictions from either models, samples from the union of the two error regions contribute to ambiguity. Hence, Model 1 and 2 bear high predictive multiplicity despite being group-fair. 

In the lower right panel, Model 1 and 2 attain low fairness and low predictive multiplicity. Both models have higher accuracy for Group 2 than Group 1, so they are both unfair. The error regions completely overlap, which means that the two models are making the same error---ambiguity is 0. 
\end{example}

The schematic diagram shows predictive multiplicity is not captured by OAE. Building on this idea, we show that, for any data distribution, it is always possible to construct classifiers with very stringent accuracy and fairness constraint but maximal ambiguity.

\begin{proposition}[Existence of Rashomon Set with Worst Case Ambiguity]
    For any dataset distribution $\calD$, empirical loss $\epsilon \in [0,0.5]$ and number of models $m>0$, assuming no constraint on the hypothesis class $\calH$, there exists an empirical Rashomon set $\hatRm\in \hatR$ satisfying Overall Accuracy Equality perfectly but with maximal ambiguity, i.e., $\alpha_{\hatRm}= \min\{100\%,m\epsilon\}$.
    \label{prop::worst case}
\end{proposition}

Proposition \ref{prop::worst case} implies that with a large enough set of competing models ($m>\frac{1}{\epsilon}$), 100\% of the samples in the dataset can receive conflicting predictions from this set of perfectly fair models. 

In the next example, we demonstrate that, counter-intuitively, adding a fairness constraint can enlarge the set of optimal models, thereby increasing predictive multiplicity. This points to a fundamental reason why adding fairness constraints can lead to more arbitrariness in model decisions.

\begin{figure}[!tb]
\centering
\includegraphics[width=1\textwidth]{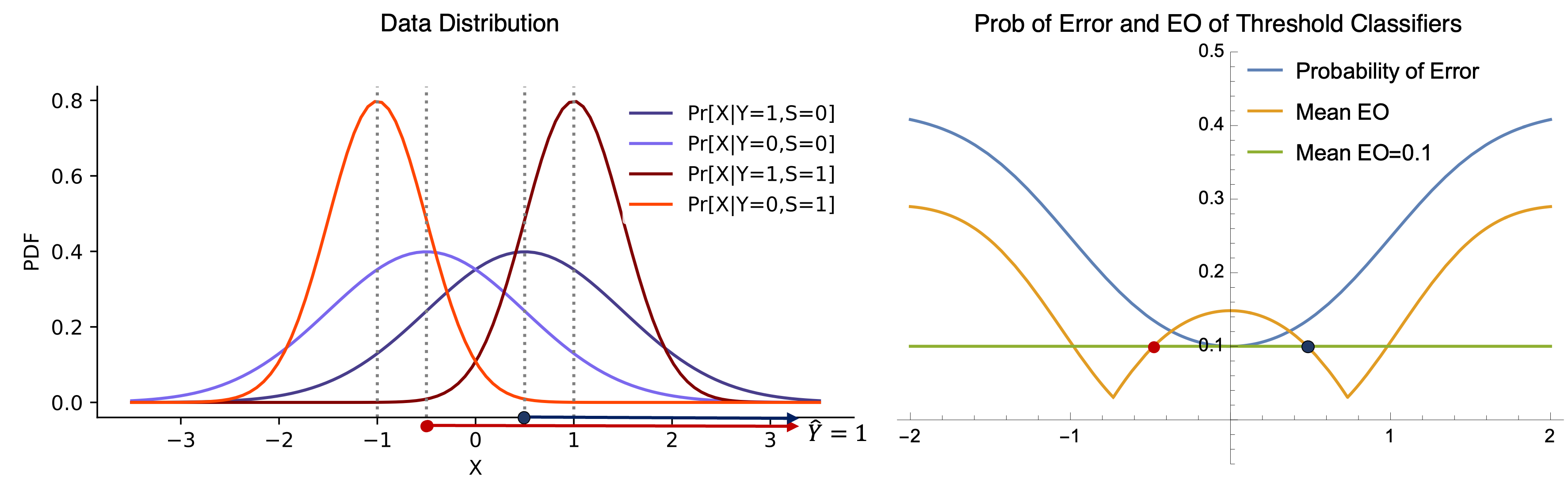}
\caption{\footnotesize
Data distribution of a population with two groups used in Example 2 (\textbf{Left}). In \textbf{Right}, without the \textrm{Mean EO} constraint~\eqref{eq::EO} (green line), there is a unique optimal classifier (with threshold 0) that attains the smallest probability of error (blue line). Adding the \textrm{Mean EO} constraint enlarges the set of optimal threshold classifiers to two classifiers (red and blue dots) with  indistinguishable accuracy and fairness levels (\textbf{Right}) but different decision regions. We illustrate the decision regions of each classifier as red and blue arrows on the \textbf{Left}. 
}
\label{fig::normal example}
\end{figure}

\begin{example}[Arbitrariness of Threshold Classifiers with Fairness Constraint] Given a data distribution of a population with two groups (Figure \ref{fig::normal example} \textbf{Left}), consider the task of selecting a threshold classifier that predicts the true label. Without fairness considerations, the optimal threshold is 0 -- i.e., assigning positive predictions to samples with $\bx>0$ and negative predictions to $\bx \leq 0$ minimizes the probability of error (Figure \ref{fig::normal example} \textbf{Right}). This optimal model is unique. Adding a fairness constraint that requires $\textrm{Mean EO}\leq 0.1$, the previously optimal classifier at 0 (with $\textrm{Mean EO}=0.15$, \textbf{Right}) does not meet the fairness criteria. Searching over the threshold classifiers that minimize the probability of error while satisfying \textrm{Mean EO} constraint yields two equally optimal models (red and blue dots \textbf{Right}) with distinct decision regions (red and blue arrows \textbf{Left}). 
Even in this simple hypothesis class, the addition of fairness constraints yields multiple models with indistinguishable fairness and accuracy but with distinct decision regions. The arbitrary selection between these points can lead to arbitrary outputs to points near the boundary.

\label{example 2}
\end{example}

\section{Ensemble Algorithm for Arbitrariness Reduction}\label{sec:ensemble} 

To tackle the potential arbitrariness cost of fairness intervention algorithms, we present a disparity-reduction mechanism through ensembling. We provide theoretical guarantees and numerical benchmarks to demonstrate that this method significantly reduces the predictive multiplicity of fair and accurate models. 

In a nutshell, given competing models $h_1,\cdots,h_m$, we argue that the disparity in their score assignment can be reduced by considering a convex combination of them, defined as follows. 

\begin{defn}[Ensemble Classifier] \label{def:ensemble}
    Given $m$ classifiers $\bh = \{ h_1,\cdots,h_m:\BR^d\to [0,1]\}$ and a vector $\blambda \in \bDelta_m$, we define the \emph{$\blambda$-ensembling} of the $h_i$ to be the convex combination
    \begin{equation}
        \bh^{\ens,\blambda}\defined \sum_{i\in [m]} \lambda_i h_i.
    \end{equation}
\end{defn}

\subsection{Concentration of Ensembled Scores}

We prove in the following result that \emph{any} two different ensembling methods agree for fixed individuals with high probability. Recall that we fix a dataset $\calD$ and a set of competing models $\calT(\calD)$ coming from a stochastic training algorithm $\calT$ (see Section~\ref{sec:setup}). All proofs are provided in Appendix \ref{sec:: proofs}.

\begin{theorem}[Concentration of Ensembles' Scores] \label{Thm:concentration}

Let $h_1,\hdots,h_m;\tilde{h}_1,\hdots,\tilde{h}_m \overset{iid}{\sim} \mathrm{Unif}(\calT(\calD))$ be $2m$ models drawn from $\calT(\calD)$, and $\bh^{\ens,\blambda},\tilde{\bh}^{\ens,\bgamma}$ be the ensembled models for $\blambda,\bgamma \in \bDelta_m$ (see Definition~\ref{def:ensemble}) satisfying $\|\blambda\|_2^2,\|\bgamma\|_2^2\le c/m$ for an absolute constant $c$. 

For every $\bx\in \BR^d$ and $\nu \ge 0$, we have the exponentially-decaying (in $m$) bound
\begin{equation}\label{eq:score_variation_prob}
    \BP\left( \left| \bh^{\textup{ens},\blambda}(\bx) - \tilde{\bh}^{\textup{ens},\bgamma}(\bx) \right| \ge \nu \right) \le 4 e^{-\nu^2m/(2c)}.
\end{equation}

In particular, for any validation set $\calD_{\textup{valid.}}\subset \BR^d$ of size $|\calD_{\textup{valid.}}|=n$, we have the uniform bound

\begin{equation}
    \BP\left( \left| \bh^{\textup{ens},\blambda}(\bx) - \tilde{\bh}^{\textup{ens},\bgamma}(\bx) \right| < \nu \ \text{ for all } \bx\in \calD_{\textup{valid.}}\right) > 1 - 4 n e^{-\nu^2 m /(2c)}.
\end{equation}
\end{theorem}

\subsection{Concentration of Predictions Under Ensembling}

The above theorem implies that we can have a dataset of size that is exponential in the number of accessible competing models and still obtain similar scoring for \emph{any} two ensembled models (uniformly across the dataset). 

In practice, one cares more about the agreement of the final prediction of the classifiers. The following result extends Theorem~\ref{Thm:concentration} to the concentration of thresholded classifiers. For this, we need to define the notion of \emph{confident classifiers}.

\begin{defn}[Confident Classifier] \label{def:strong}
    For a binary classifier $h:\BR^d\to [0,1]$, a probability measure $P_{\bX}$ over $\BR^d$, and constants $\delta,\theta\in [0,1]$, we say that $h$ is a \emph{$(P_{\bX},\delta,\theta)$-confident  classifier} if it holds that 
    \begin{equation}
        \BP\left( \left| h(\bX) - \frac12 \right| < \delta \right) < \theta.
    \end{equation}
\end{defn}
In other words, $h$ is a confident classifier if it is ``more sure'' of its predictions. 
We observe in experiments that models corrected by fairness interventions have scores concentrated around 0 and 1.

Using confident classifiers, we are able to extend Theorem~\ref{Thm:concentration} to thresholded ensembles, as follows. 


\begin{theorem} \label{thm::score variation}
    Let $\bh^{\ens,\blambda},\tilde{\bh}^{\ens,\bgamma}$ be as in Theorem~\ref{Thm:concentration}, and assume that both ensembled classifiers are $(P_{\bX},\delta,\theta)$-confident in the sense of Definition~\ref{def:strong}. Let $f(t)\defined \indicator[t\ge 0.5]$ be the thresholding function. For any set $\calD_0 \subset \BR^d$ of size $|\calD_0|=n_0$, we may guarantee the probability of agreement in the predictions for all samples under the two ensembles to be at least
    \begin{equation}
        \BP\left( f(\bh^{\ens,\blambda}(\bx)) = f(\tilde{\bh}^{\ens,\bgamma}(\bx)) \ \text{ for every } \bx\in \calD_0\right) \ge 1 - \left( 4 e^{-2\delta^2 m/c  }+2\theta \right)n_0.
    \end{equation}
\end{theorem}
We note that in the fairness-intervention setting, the set $\calD_0$ in the above theorem would be chosen as the subset of samples having the same group attribute. Thus, the size $n_0$ of $\calD_0$ would be significantly smaller than the total size of the dataset, and the parameter $\theta$ then can be required to be moderately small. 
\begin{remark}
    In Appendix \ref{sec:: optimize param}, we discuss how to optimize the ensembling parameters $\blambda$. In the next section, we will stick to the uniform ensembling: $\bh^{\ens,\blambda} = \frac{1}{m}\sum_{j \in [m]} h_j$, i.e., $\blambda = \frac{1}{m}\mathbf{1}$. This simple uniform ensemble suffices to illustrate the main goal of this paper: that arbitrariness can be a by-product of fairness intervention methods, and ensembling can mitigate this unwanted effect.
\end{remark}

\section{Experimental Results}\label{sec:exp}
We present empirical results to show that arbitrariness is masked by favorable group-fairness and accuracy metrics for multiple fairness intervention methods, baseline models, and datasets. We also demonstrate the effectiveness of the ensemble in reducing the predictive multiplicity of fair models. 

\paragraph{Setup and Metrics.}
We consider three baseline classifiers (\textsc{Base}): random forest (RF), gradient boosting (GBM), and logistic regression (LR), implemented by Scikit-learn \citep{pedregosa2011scikit}. By varying the random seed, we obtain 10 baseline models with comparable performance. Then, we apply various state-of-the-art fairness methods (details in Appendix \ref{sec:: discussion fairness}) on the baseline models to get competing fair models.

On the test set, we compute mean accuracy, \textrm{Mean EO} (Definition \ref{eq::EO}), and predictive multiplicity levels on competing models before and after fairness interventions. We use ambiguity (Definition \ref{def:: ambiguity}) and score standard deviations (Definition \ref{def:: score SD}) as metrics for predictive multiplicity.

\paragraph{Datasets.}
We report predictive multiplicity and benchmark the ensemble method on three datasets. We use two datasets in the education domain: the high-school longitudinal study (HSLS) dataset \citep{ingels2011high,jeong2022fairness} and the ENEM dataset
 \citep{cury2022instituto} (see \citet{alghamdi2022beyond} Appendix B.1). The ENEM dataset contains Brazilian college entrance exam scores along with student demographic information and socio-economic questionnaire answers (e.g. if they own a computer). After pre-processing, the dataset contains ~1.4 million samples with 139 features. Race is used as the group attribute $S$, and Humanities exam score is used as the label $Y$. Scores are quantized into two classes for binary classification. The race feature $S$ is binarized into White and Asian ($S = 1$) and others ($S = 0$). The experiments are run with a smaller version of the dataset with 50k samples. For completeness, we also report results on UCI Adult \citep{lichman2013uci} in Appendix \ref{sec:: additional experiments}.
\begin{figure}[!tb]
\centering
\includegraphics[width=1.0\textwidth]{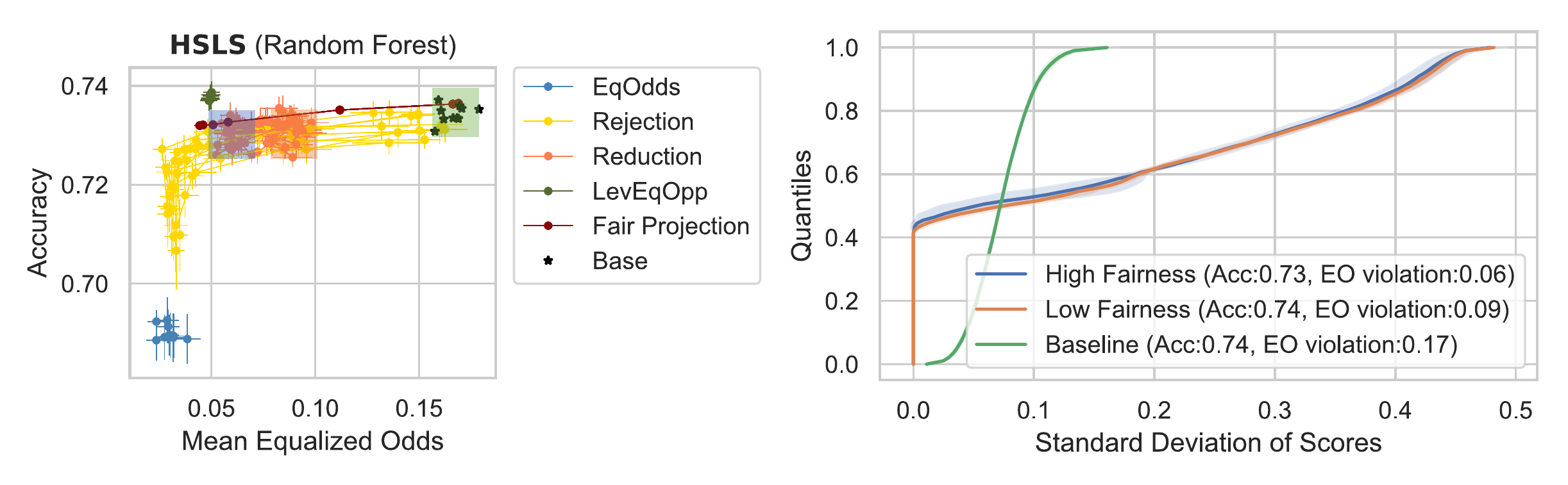}
 \caption{\footnotesize
 Quantile plot on score std. for low-fairness, high-fairness, and baseline bins on HSLS. Accuracy-fairness frontier does not reveal arbitrariness in competing models. 
\textbf{Left}: We repeat 5 fairness interventions with various random seeds on the HSLS dataset, and acquire 10 classifiers for each intervention with competing accuracy-vs-fairness levels (horizontal axis is computed using~\eqref{eq::EO}). 
\textbf{Right}: We pick three sets of classifiers with competing performance (blue, red, and green shaded rectangles \textbf{Left}) and measure the per-sample standard deviation of scores (Definition \ref{def:: score SD}). 
}
 \label{fig::hsls}
\end{figure}
\begin{figure}[!tb]
\centering
\includegraphics[width=1.0\textwidth]{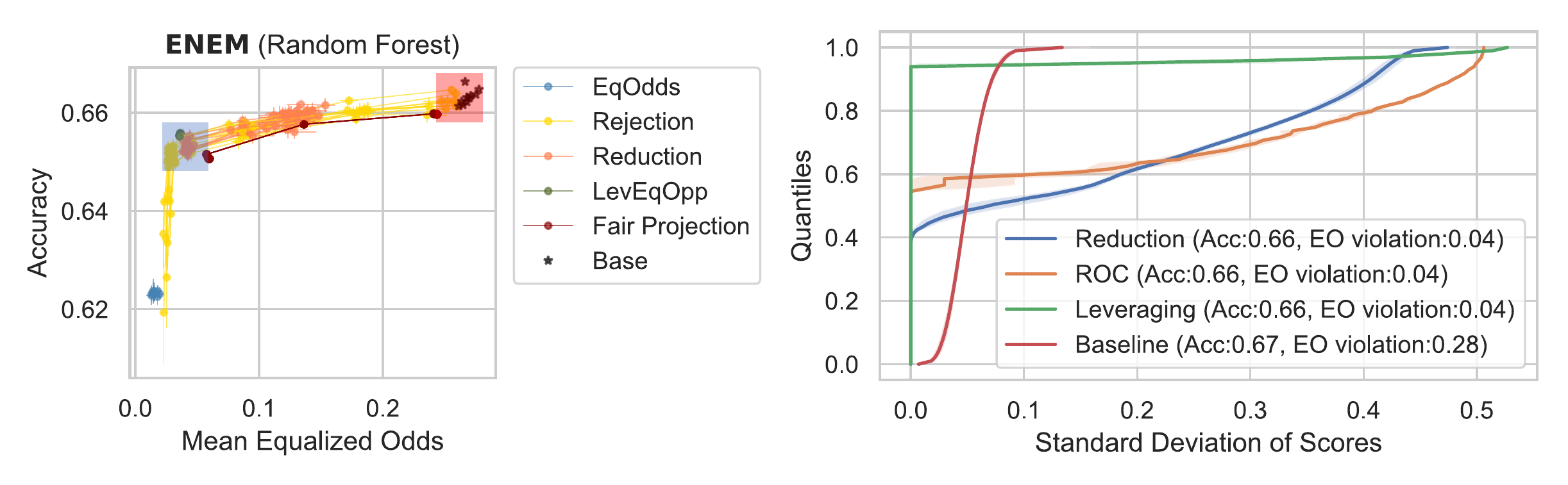}
\caption{\footnotesize
Quantile plot on high-fairness bin for various fairness interventions v.s. baseline on ENEM. \textbf{Left}: Fairness-Accuracy frontier. \textbf{Right}: Fair models produce larger score std. at top percentiles compared to the baseline model (horizontal axis computed via~\eqref{eq::EO}).  (\textsc{Rejection} and \textsc{Leveraging} output thresholded scores directly.)}
\label{fig::enem_multiple}
\end{figure}
\paragraph{Results that Reveal Arbitrariness.} 
We juxtapose the fairness-accuracy frontier and metrics for predictive ambiguity to reveal arbitrariness masked by favorable group-fairness and accuracy metrics in Figure \ref{fig::Enem quantile}, \ref{fig::hsls}, and \ref{fig::enem_multiple}. Starting with 10 baseline classifiers by varying the random seed used to initialize the training algorithm, we apply the fair interventions \textsc{Reduction} \citep{agarwal2018reductions}, \textsc{Rejection} \citep{kamiran2012decision}, \textsc{Leveraging} \citep{chzhen2019leveraging} to obtain point clouds of models with comparable fairness and accuracy metrics. 
In Figure \ref{fig::enem_multiple}, we take models that achieve very favorable accuracy and \textrm{Mean EO} metrics (in blue rectangle in \textbf{Left}) and plot the std. of scores to illustrate predictive multiplicity \textbf{Right}. Group fairness violations are greatly reduced (from 0.28 in baseline to 0.04 in fair models) at a small accuracy cost (from 67\% in baseline to 66\% in fair models). 
However, there is higher arbitrariness. 

Compared to baseline (red curve), fair models corrected by \textsc{Reduction} and \textsc{ROC} produce lower score arbitrariness for the bottom 50\% but much higher arbitrariness for the top 50\% of samples; importantly, the induced arbitrariness becomes \emph{highly nonuniform across different individuals} after applying the two fairness intervention. We observe that \textsc{Leveraging} produce models that agree on ~90\% of the samples, thereby not inducing concerns of arbitrariness.

\begin{figure}[!tb]
\centering
\includegraphics[width=0.9\textwidth]{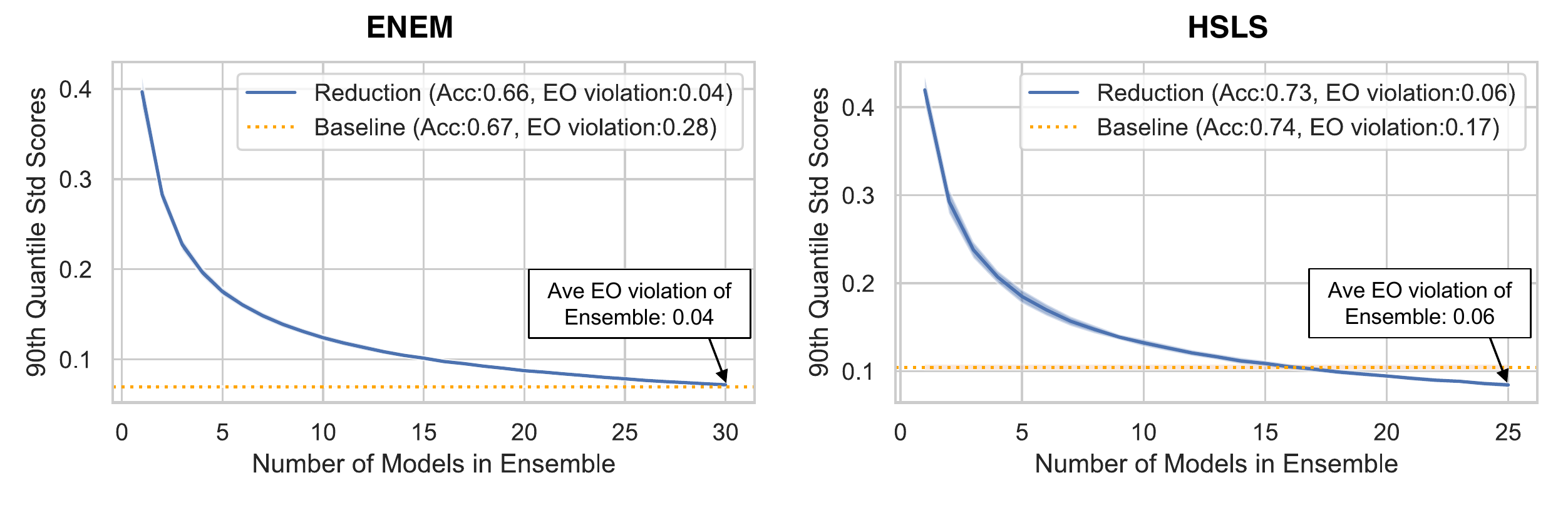}
 \caption{\footnotesize
 Standard deviation of ensembled models trained on ENEM and HSLS with baseline random forest classifiers. We fix the high-fairness bin and vary the number of models $m$ in each ensemble. As we increase the number of ensembles, score std. (on 10 ensembles) drops and meets the score std. of 10 baseline RFC when $m=30$ on ENEM and $m = 17$ on HSLS. (\textrm{Mean EO} is computed using~\eqref{eq::EO}.}
 \label{fig::ensemble convergence}
\end{figure}

Remarkably, arbitrariness does not vary significantly among models with different fairness levels. We consider two sets of models trained with high and low fairness constraints using \textsc{Reduction} in Figure \ref{fig::Enem quantile} and Figure \ref{fig::hsls}. In Figure \ref{fig::hsls}, score std. of the high-fairness models (blue curve) and the low-fairness models (orange curve) overlap, while both deviate significantly from baseline.
\paragraph{Results on the Effectiveness of Ensembling.}
We pair our proofs in Section 4 with experiments that demonstrate the concentration of scores of ensembled models. In Figure \ref{fig::ensemble convergence} \textbf{Left}, taking the competing models in the high-fairness bins corrected with \textsc{Reduction} that achieve an \textrm{Mean EO} violation of $0.04$ but very high score std. for half of the samples (blue rectangle in Figure \ref{fig::enem_multiple}), we ensemble the models with increasing number of models per ensemble ($m$) ranging from 1 to 30. For each $m$, we measure std. of scores in 10 such ensembles. The top percentile std. of the ensembled fair models drops to baseline with 30 models. Similar convergence occur on the HSLS dataset. Importantly, the ensembled models are still fair, the \textrm{Mean EO} violations of the ensembled models remain low.

\section{Final Remarks and Limitations}

We demonstrate in this paper that arbitrariness is a facet of responsible machine learning that is orthogonal to existing fairness-accuracy analyses. Specifically, fairness-vs-accuracy frontiers are insufficient for detecting arbitrariness in the predictions of group-fair models: two models can have the same fairness-accuracy curve while at the same time giving widely different predictions for subsets of individuals. We demonstrate this undesirable phenomenon both theoretically and experimentally on state-of-the-art fairness intervention methods. Furthermore, towards mitigating this arbitrariness issue, we propose an ensemble algorithm, where a convex combination of several competing models is used for decision-making instead of any of the constituent models. We prove that the scores of the ensemble classifier concentrate, and that the ensuing predictions can be made to concentrate under mild assumptions. Importantly, we exhibit via real-world experiments that our proposed ensemble algorithm can reduce arbitrariness while maintaining fairness and accuracy.

The proposed framework for estimating the predictive multiplicity of fairness interventions requires re-training multiple times, limiting its applicability to large models. 
We consider model variation due to randomness used during training. In practice, competing models may exist due to inherent uncertainty (i.e., a non-zero confidence interval) when evaluating model performance on a finite test set. In this regard, models with comparable average performance can be produced by searching over this Rashomon set even if training is deterministic (e.g., equivalent to solving a convex optimization). An interesting future direction is to explore the multiplicity cost of fairness interventions in such deterministic settings.
Our ensembling strategy may not guarantee that the ensemble classifier ensures fairness constraints due to the non-convex nature of such constraints. Though we empirically observe that fairness constraints are indeed satisfied by the ensemble model, proving such guarantees theoretically would be valuable.

\paragraph{Acknowledgements.}
The authors would like to thank Arpita Biswas and Bogdan Kulynych for helpful discussions on initial ideas. This material is based upon work supported by the National Science Foundation under grants CAREER 1845852, CIF 1900750, and FAI 2040880, and by Meta Ph.D. fellowship.


\bibliographystyle{apalike}
\bibliography{reference}

\clearpage
\section*{Appendix}
\appendix

The appendix is divided into the following four parts. Appendix \ref{sec:: proofs}: Proofs of theoretical results; Appendix \ref{sec:: optimize param}: Discussion on optimizing ensemble parameters; Appendix \ref{sec:: discussion fairness}: Additional discussions on group fairness and fairness interventions; and Appendix \ref{sec:: additional experiments}: Additional experiments and details on the experimental setup.

\section{Proofs of theoretical results}
\label{sec:: proofs}

\paragraph{Proof of Proposition 3.1}
\begin{proof}
    
Given a dataset $\calD \triangleq \{(\bx_i,s_i,y_i)\}_{i=1}^n$ sampled from any distribution and parameters that define the empirical Rashomon set $\hatRm\in \hatR$ -- empirical loss $\epsilon\in [0,0.5]$, number of models $m>0$, a hypothesis class $\calH$ with no constraint, we can construct the empirical Rashomon set $\hatRm$ in a way that the models satisfy Overall Accuracy Equality (OAE from Definition \ref{def::OAE}) perfectly but with maximal ambiguity, i.e. $\alpha_{\hatRm}=\min\{100\%,m\epsilon\}$. 

Here is the main idea behind the construction: for each of the $K$ groups $\calS = [K]$, OAE requires that each model gets exactly $\lfloor n_j\epsilon \rfloor$ samples wrong, where $n_j$ denotes the number of samples in group $j\in[K]$. The floor operator is taken to get an integer number of samples. To maximize ambiguity, the error regions of the models are made as disjoint as possible (recall example in Figure \ref{fig::diagram}). 

Define $\hatRm = \{h_1,...,h_m\}$ as follows: for every group $j \in\calS$, start with samples that belong to group $j$, i.e., $\calD_j\triangleq \{(\bx_i,s_i,y_i) \mid s_i = j\}_{i=1}^{n_j}$. Partition $\calD_j$ into subsets $\{\calA_u\}_{u=1}^{\lfloor 1/\epsilon \rfloor+1}$, each of size $n_j\epsilon$ except for the last subset. The last subset takes the remainder of the samples, i.e., $\calD_j \setminus  (\calA_1 \cup ... \cup \calA_{\lfloor 1/\epsilon \rfloor})$. 

Case 1: Consider the case where $m >\frac{1}{\epsilon}$. Define the first $\lfloor 1/\epsilon \rfloor +1$ models $h_1,...,h_{\lfloor 1/\epsilon \rfloor+1}$ as follows:
\begin{align}
    h_u(x_i) = 
    \begin{cases} 
      1-y_i & \forall x_i\in \calA_u \\
      y_i & \text{o.w}
   \end{cases}
\end{align}

By construction so far, the first $h_1,...,h_{\lfloor 1/\epsilon \rfloor+1}$ models' error regions are disjoint since $\calA_1,...,\calA_{\lfloor 1/\epsilon \rfloor+1}$ forms a partition of $\calD_j$. Add arbitrary samples to $h_{\lfloor 1/\epsilon \rfloor+1}$'s error region so that its loss is $\lfloor n_j\epsilon\rfloor$. For each model in the remaining $m$ models of $\hatRm$, take any $\lfloor n_j\epsilon\rfloor$ samples from $\calD_j$ to be the error region. All models have error rate $\lfloor n_j\epsilon\rfloor$ for group $j$. Repeat this process for every group, so the models achieve OAE, the same accuracy across groups.

The ambiguity of this $\hatRm$ is 100\% since for all $(\bx_i,s_i,y_i)\in \calD$, the distinct error regions defined above ensure that there are conflicting predictions between 2 models in $\hatRm$.

Case 2: Consider the case where $m \leq \frac{1}{\epsilon}$. For every group $j\in [K]$, partition $\calD_j$ similarly and take the first $m$ subsets $\calA_1,...,\calA_m$ of size $\lfloor n_j\epsilon\rfloor$ to be the error regions of the $m$ models. Hence, we again achieve an error rate of $\lfloor n_j\epsilon\rfloor$. Repeat for all groups and the models satisfy OAE.

The ambiguity of this $\hatRm$ is $m\epsilon$, which is the maximal disagreement the models can achieve.

A caveat about $\calD$ is that each group cannot have too few samples, otherwise ensuring OAE is infeasible. In the extreme case when there is only a sample for a group, accuracy can either be 0\% or 100\%, which makes it infeasible to ensure OAE or other group fairness metrics that boils down to equalizing performance across groups.
\end{proof}

\paragraph{Proof of Theorem 4.1}

\begin{proof}
    We assume that $\|\blambda\|_2^2, \|\bgamma\|_2^2\le c/m$ for an absolute constant $c$, e.g., we have $c=1$ for the uniform ensembling $\blambda=(1/m,\cdots,1/m)$ as then $\|\blambda\|_2^2=1/m$. Fix $\bx$, and denote the mean of the classifiers $\mu_{\bx}=\BE_{h\sim \calT(\calD)}[h(\bx)]$. The mapping $(h_1,\cdots,h_m)\mapsto \bh^{\ens,\blambda}$ satisfies the bounded-difference condition in the McDiarmid inequality. Indeed, changing $h_i$ can change $\bh^{\ens,\blambda}$ by at most $\lambda_i$. Furthermore, $\bh^{\ens,\blambda}(\bx)$ has the mean
    \begin{equation}
        \BE\left[ \bh^{\ens, \blambda}(\bx) \right] = \sum_{i\in [m]} \lambda_i \BE[h_i(\bx)] = \mu_{\bx} \sum_{i\in [m]} \lambda_i = \mu_{\bx}. 
    \end{equation}
    Hence, by Mcdiarmid's inequality, we have the bound
    \begin{equation}
        \BP\left( \left| \bh^{\ens,\blambda}(\bx) - \mu_{\bx} \right| \ge \nu \right) \le 2 \exp\left( \frac{-2\nu^2}{\sum_{i\in [m]} \lambda_i^2} \right) \le  2 \exp\left( \frac{-2\nu^2m}{c} \right).
    \end{equation}
    The same inequality holds for $\bgamma$ in place of $\blambda$: 
    \begin{equation}
        \BP\left( \left| \tilde \bh^{\ens,\bgamma}(\bx) - \mu_{\bx} \right| \ge \nu \right)  \le  2 \exp\left( \frac{-2\nu^2m}{c} \right).
    \end{equation}
    Therefore, we obtain the bound
    \begin{align}
        1- \BP\left( \left| \bh^{\ens,\blambda}(\bx) -  \tilde\bh^{\ens,\bgamma}(\bx) \right| \ge \nu \right) &= \BP\left( \left| \bh^{\ens,\blambda}(\bx) - \mu_{\bx} + \mu_{\bx} -  \tilde\bh^{\ens,\bgamma}(\bx) \right| < \nu \right) \\
        &\ge \BP\left( \left| \bh^{\ens,\blambda}(\bx) - \mu_{\bx}\right| + \left|  \tilde\bh^{\ens,\bgamma}(\bx) - \mu_{\bx} \right| < \nu \right) \\
        &\ge \BP\left( \left| \bh^{\ens,\blambda}(\bx) - \mu_{\bx}\right| < \frac{\nu}{2} \cap  \left|  \tilde\bh^{\ens,\bgamma}(\bx) - \mu_{\bx} \right| < \frac{\nu}{2} \right) \\
        &= 1-\BP\left( \left| \bh^{\ens,\blambda}(\bx) - \mu_{\bx}\right| \ge \frac{\nu}{2} \cup  \left|  \tilde\bh^{\ens,\bgamma}(\bx) - \mu_{\bx} \right| \ge \frac{\nu}{2} \right) \\
        &\ge 1 - 4 \exp\left( \frac{-\nu^2m}{2c} \right),
    \end{align}
    where the first inequality comes from triangle inequality, the following from probability of subset of events ($\mathbb{P}(A)\geq \mathbb{P}(B)$ if $A\supseteq B$), the equality from taking complement, and the last line from applying McDiarmid's inequality along with the union bound. 

    Finally, applying the union bound on $\calD_{\text{valid.}}$ with $|\calD_{\text{valid.}}|=n$, we obtain the bound
    \begin{align}
        \BP\left( \bigcap_{\bx \in \calD_{\text{valid.}}} \left| \bh^{\textup{ens},\blambda}(\bx) - \tilde{\bh}^{\textup{ens},\bgamma}(\bx) \right| < \nu \right) &= 1- \BP\left( \bigcup_{\bx \in \calD_{\text{valid.}}} \left| \bh^{\textup{ens},\blambda}(\bx) - \tilde{\bh}^{\textup{ens},\bgamma}(\bx) \right| \ge \nu \right) \\
        &\ge 1 - 4 n  \exp\left( \frac{-\nu^2m}{2c} \right),
    \end{align}
    and the proof is complete.
\end{proof}

\paragraph{Proof of Theorem 4.2}
\begin{proof}
    The main idea is as follows: first observe that for the ensembled labels to disagree on a sample (given that the scores are bounded away from $\frac{1}{2}$ with high probability), the two models need to produce scores in the range $[0,\frac{1}{2}-\delta]\cup[\frac{1}{2}+\delta,1]$. This means that the scores need to deviate at least $2\delta$ which has an exponentially low probability given Theorem 4.1.
    
    
    We will show that
    \begin{equation}
        \BP\left( \bigcup_{\bx \in\calD_0} f( \bh^{\ens, \blambda}(\bx)) \neq f(\tilde{\bh}^{\ens,\bgamma}(\bx)) \right) \le \left( 4 e^{-2\delta^2 m/c}+2 \theta \right)n_0.
    \end{equation}
    Indeed, for each fixed $\bx\in \calD_0$, we may reduce the failure probability to the case of separation of scores:
    \begin{align}
        \BP\left(  f( \bh^{\ens, \blambda}(\bx)) \neq f(\tilde{\bh}^{\ens,\bgamma}(\bx))  \right) &\le \BP{\Bigg (} \left( \bh^{\ens,\blambda}(\bx) \in \left[ 0, \frac12 - \delta \right] \cap  \tilde\bh^{\ens, \bgamma}(\bx) \in \left[ \frac12+\delta, 1 \right] \right) \\
        &\qquad \cup \left(  \tilde\bh^{\ens,\bgamma}(\bx) \in \left[ 0, \frac12 - \delta \right] \cap \bh^{\ens, \blambda}(\bx) \in \left[ \frac12+\delta, 1 \right] \right) \\
        &\qquad \cup \bh^{\ens,\blambda}(\bx) \in \left[ \frac12-\delta,\frac12+\delta\right] \\
        &\qquad \cup  \tilde\bh^{\ens,\bgamma}(\bx) \in \left[ \frac12-\delta,\frac12+\delta\right] {\Bigg )} \\
        &\le \BP\left( \left| \bh^{\ens, \blambda}(\bx) -  \tilde\bh^{\ens, \bgamma}(\bx) \right| \ge 2\delta \right) + 2\theta \\
        &\le 4 \exp\left( \frac{-2\delta^2m}{c}  \right) + 2\theta.
    \end{align}
    Finally, applying the union bound, we obtain that
    \begin{align}
        \BP\left( \bigcap_{\bx \in\calD_0} f( \bh^{\ens, \blambda}(\bx)) = f(\tilde{\bh}^{\ens,\bgamma}(\bx)) \right) &= 1- \BP\left( \bigcup_{\bx \in\calD_0} f( \bh^{\ens, \blambda}(\bx)) \neq f(\tilde{\bh}^{\ens,\bgamma}(\bx)) \right) \\
        &\ge 1- \left( 4 e^{-2\delta^2 m/c}+2 \theta \right)n_0,
    \end{align}
    and the proof is complete.
\end{proof}

\section{Discussion on optimizing ensemble parameters}
\label{sec:: optimize param}

We have taken the weights $\blambda \in \bDelta_m$ which determines the ensembled model $\bh^{\ens,\blambda}$ to be fixed. We explain here how $\blambda$ can be optimized according to a given cost. Specifically, given a loss function $\ell : [0,1]\times \{0,1\}\to \BR_+$, we can search for the optimal $\blambda \in \bDelta_m$ that minimizes the total cost
\begin{equation} \label{eq:SLT population}
    L_{\ens}(\blambda) \defined \BE\left[ \ell(\bh^{\ens,\blambda}(X),Y\right) ].
\end{equation}
For the above optimization problem, we think of the constituent models $h_1,\cdots,h_m$ as being fixed and the randomness is from that of $(X,Y)$. 

However, in practice, we have access to only samples $(\bx_i,y_i) \sim P_{\bX,Y}$. Thus, we consider minimizing the regularized sample mean (for fixed $\beta>0$)
\begin{equation} \label{eq:SLT sample}
    \hat{L}_{\ens}(\blambda) \defined \frac{1}{n} \sum_{i\in [n]} \ell\left( \bh^{\ens,\blambda}(\bx_i), y_i \right) + \frac{\beta}{\sqrt{n}} \|\blambda\|_2^2.
\end{equation}
The $2$-norm regularization is added to facilitate proving convergence. 
This convergence result can be obtained via known results from statistical learning theory, e.g., using Theorem~13.2 in~\citep{hajek2019statistical}. Specifically, consider the following two restrictions:
\begin{itemize}
    \item Consider only $\blambda\in \bDelta_m$ satisfying $\|\blambda\|_2\le \alpha $ for a prescribed $\alpha$. Note that we may take $\alpha=1$ to encapsulate the whole probability simplex. However, we may choose $\blambda$ to be a slight modification of the uniform ensembling, in which case we would have $\alpha$ of order $1/\sqrt{m}$. 
    \item Assume that the function $\blambda \mapsto \ell(\bh^{\ens,\blambda}(\bx),y)$ is convex and $A$-Lipschitz for each fixed $(\bx,y)$.
\end{itemize}
In this case, choosing $\beta = A/\alpha$ and denoting the optimizers
\begin{equation}
    \blambda^{(n)} \defined \underset{\|\blambda\|_2\le \alpha}{\textup{argmin}} \ \hat{L}_{\ens}(\blambda),
\end{equation}
we can bound the utility of these minimizers by
\begin{equation}
    \BP\left( L_{\ens}(\blambda^{(n)}) \le \inf_{\|\blambda\|_2\le \alpha} \ L_{\ens}(\blambda) + \frac{\beta\alpha^2}{\sqrt{n}} \cdot \left( 1 + \frac{1}{\delta} + \frac{8}{\delta \sqrt{n}} \right) \right) \ge 1-\delta
\end{equation}
for any $\delta\in (0,1)$. 


\section{Additional discussion on group fairness and fairness interventions}
\label{sec:: discussion fairness}
In addition to OAE (Definition \ref{def::OAE}), two other important fairness criteria are Statistical Parity \citep{dwork2015preserving} and Equalized Odds \citep{hardt2016equality}.
\begin{defn}[SP]
    $\Pr(\hat{Y} = 1 | S = s) = \Pr(\hat{Y} = 1 | S = s')$ for all groups $s,s'\in [K]$ . 
\end{defn}
\begin{defn}[EO]
    $\Pr(\hat{Y} = 1 | S = s, Y = b) = \Pr(\hat{Y} = 1 | S = s', Y=y)$ for all groups $s,s'\in [K]$, and binary labels $y\in \{0,1\}.$ 
\end{defn}

Essentially, SP requires the predicted label $\hat{Y} \triangleq \arg\max h(\bX)$ to be independent of the group membership $S$~\citep{10.1145/2090236.2090255}. In comparison, EO conditions on both group and the true label~\citep{hardt2016equality}. 
EO improves upon SP in the sense that it does not rule out the perfect classifiers whenever the true label $Y$ is correlated with the group membership $S$~\citep{agarwal2018reductions}. In practice, we quantify EO violation by measuring Mean EO as in Equation \ref{eq::EO} (for two groups) and, more generally, in Equation \ref{eq::EO violation multi} below (beyond two groups). Similarly, we can measure SP violation as in Equation \ref{eq::SP violation}.
\begin{equation} \label{eq::EO violation multi}
    \textsc{Mean EO} \defined \max_{s,s' \in [K]}\frac{1}{2}\left( |\textsc{TPR}_{S=s}-\textsc{TPR}_{S=s'}| +|\textsc{FPR}_{S=s}-\textsc{FPR}_{S=s'}| \right).
\end{equation}
\begin{equation} \label{eq::SP violation}
    \textsc{SP Violation} \defined \max_{s,s' \in [K]} \frac{1}{2}\left( |\Pr(\hat{Y} = 1 | S = s) -\Pr(\hat{Y} = 1 | S = s')| \right).
\end{equation}

Next, we offer a brief discussion of various intervention mechanisms used in this paper. The fairness interventions can be categorized into two categories -- in-processing and post-processing. In-processing mechanisms incorporate fairness constraints during training. It usually add the fairness constraint to the loss function and outputs a fair classifier. Post-processing mechanisms treat the model as a black box and update its predictions to achieve the desirable fairness constraints \citep{caton2020fairness}. 

\textsc{Reduction}\citep{agarwal2018reductions}, short for exponentiated gradient reduction, is an in-processing technique that reduces fair classification to a sequence of cost-sensitive classification problems, and yields a randomized classifier with the lowest empirical error subject to the desired constraints. This technique achieves fairness with a minimal decrease in accuracy, but it is computationally expensive since it requires re-training multiple models.

\textsc{Reject option classifier}\citep{kamiran2012decision} is a postprocessing technique that achieves fairness constraints by modifying outcomes of samples in a confidence band of the decision boundary with the highest uncertainty. It gives favorable outcomes to unprivileged groups and unfavorable outcomes to privileged groups. It outputs a thresholded prediction rather than a probability over the binary labels.

\textsc{EqOdds}\citep{hardt2016equality} is a post-processing technique that formulates empirical risk minimization with fairness constraint as a linear program and modifies predictions according to the derived probabilities to achieve equalized odds.

\textsc{Fair Projection} \citep{alghamdi2022beyond} is a post-processing technique that can accommodate fairness constraints in a setting with multiple labels and multiple groups. The fair model is obtained from 'projecting' a pre-trained (and potentially unfair) classifier onto the set of models that satisfy target group-fairness requirements.

\section{Additional experiments and details on the experimental setup}
\label{sec:: additional experiments}
\begin{figure}[!tb]
\centering
\includegraphics[width=1.0\textwidth]{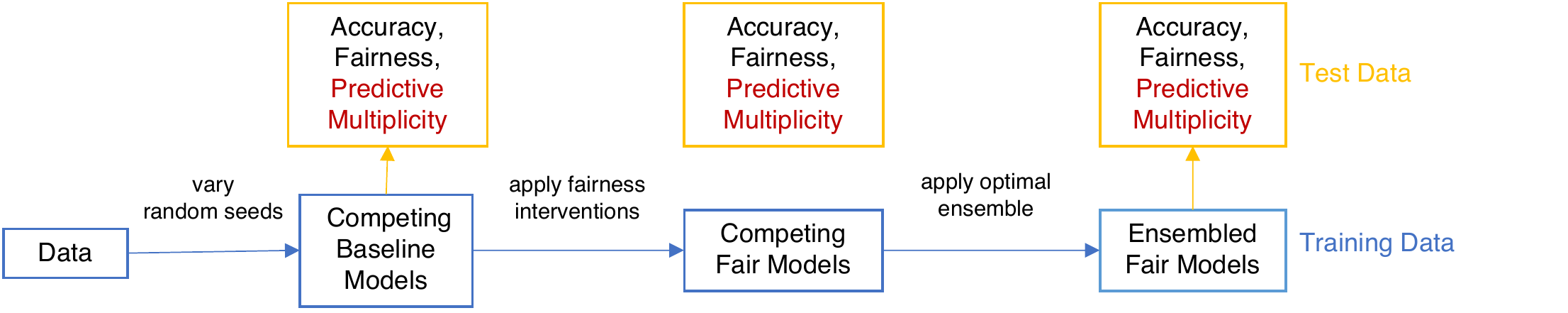}
 \caption{Flow chart of experimental procedure.}
 \label{fig::flowchart}
\end{figure}

Our proposed methodology can be summarized in the pipeline in Figure \ref{fig::flowchart}.

\subsection{Data}
The HSLS dataset\citep{ingels2011high,jeong2022fairness} is an education dataset collected from 23,000+ students across high schools in the USA. Features of the dataset contain extensive information on students' demographic information, their parents' income and education level, schools' information, and students' academic performances across years. We apply the pre-processing techniques adopted in \citet{alghamdi2022beyond}, with the number of samples reduced to 14,509. For the binary classification task with fairness constraints, the group attribute chosen is $\textsc{Race}\in \{\textsc{White},\textsc{Non-White}\}$  and the prediction label is students 9th-Grade $\textsc{GradeBin}\in\{0,1\}$, binarized according to whether a student's grade is higher or lower than the median. 

The ENEM dataset\citep{cury2022instituto} is a Brazilian high school national exam dataset introduced by \citet{alghamdi2022beyond}. It has 138 features containing students' demographic information, socio-economic questionnaire answers (e.g., parents' education level and if they own a computer), and students' exam scores. Adopting the preprocessing technique in \citet{alghamdi2022beyond}, we sample 50K samples without replacement from the processed ENEM Year 2020 data. Identical to HSLS, the group attribute chosen is $\textsc{Race}\in \{\textsc{White},\textsc{Non-White}\}$  and the prediction label is students Grade binarized into  $\textsc{GradeBin}\in\{0,1\}$ according to whether a student's grade is higher or lower than the median. 

For the widely known Adult dataset\citep{lichman2013uci}, also known as "Census Income" dataset, we choose the group attribute as  $\textsc{Sex}\in \{\textsc{Male},\textsc{Female}\}$ and predicted label to be $\textsc{Income}\in \{0,1\}$, where income bin denotes whether a person's income is higher or lower than 50K/yr.

\subsection{Competing Baseline Models}
We use the Scikit-learn implementation of logistic regression, gradient boosting, and random forest as baseline models. For logistic regression and gradient boosting, the default hyperparameter is used; for random forest, we set the number of trees and minimum number of samples per leaf to 10 to prevent over-fitting. To get 10 competing models for each hypothesis class, we use 10 random seeds (specifically 33-42).

In practice, the competing models, i.e. $h\in\hatRm$ can be obtained using different methodologies, such as sampling and adversarial weight perturbation\citep{hsu2022rashomon,watson2022predictive}. We suggest one method for sampling. First, split the data into training, validation, and test dataset. We train a set of models by changing the randomized procedures in the training process -- e.g., using different initializations, different cuts for cross-validation, data shuffling, etc. In this paper, we change the random seed feed into the baseline models to obtain competing models. We use the validation set to measure $\epsilon$ corresponding to this empirical Rashomon Set.

\subsection{Competing Fair Models}\label{app:competing-models}
For \textsc{EqOdds}, \textsc{Rejection}, and \textsc{Reduction}, we use the functions \textsc{EqOddsPostprocessing}, \textsc{RejectOptionClassification}, and \textsc{ExponentiatedGradientReduction} from AIF360 toolkits \citep{bellamy2019ai}. For \textsc{Leveraging} and \textsc{Fair Projection}, we use the codes provided in the corresponding Github repositories of \citet{chzhen2019leveraging} and \citet{alghamdi2022beyond}.

\subsection{Complete Experimental Plots}
In order to evaluate the predictive multiplicity of models with similar accuracy and fairness levels, we divide the accuracy-fairness frontier plots into 8x8 grids and put models in the corresponding bins. To compare the arbitrariness of models satisfying high/low fairness constraints, we select bins in two different MEO ranges and bins with baseline models. Then, we compute the standard deviation of scores of models corrected by \textsc{Reduction} in the three bins (high fairness/low fairness/baseline) and plot the quantile curves. 

Across three baseline model classes (random forest, gradient boosting, and logistic regression), fair models exhibit higher score arbitrariness. Especially at top quantiles, all fair models have standard deviations in scores going up to 0.5 (Figure 2,3,4). This means that for the individuals at the top percentile, the model prediction can flip if another random seed is used in model training. 

\begin{figure}[!tb]
\centering
\includegraphics[width=1.0\textwidth]{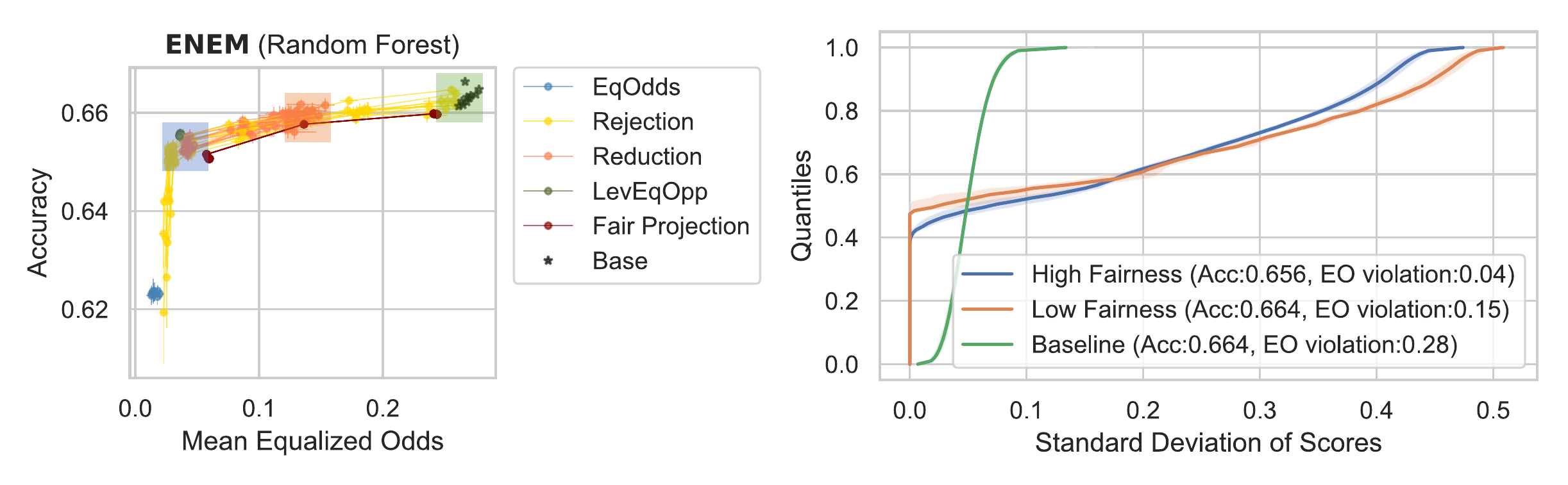}
\caption{\footnotesize \textbf{Left}: Accuracy-fairness curves of baseline random forest models v.s. fair models on the ENEM dataset.
\textbf{Right}: Quantiles of per-sample score std. across high/low fairness models and baseline.}
\label{fig::EnemRFHighLow}
\end{figure}

\begin{figure}[!tb]
\centering
\includegraphics[width=1.0\textwidth]{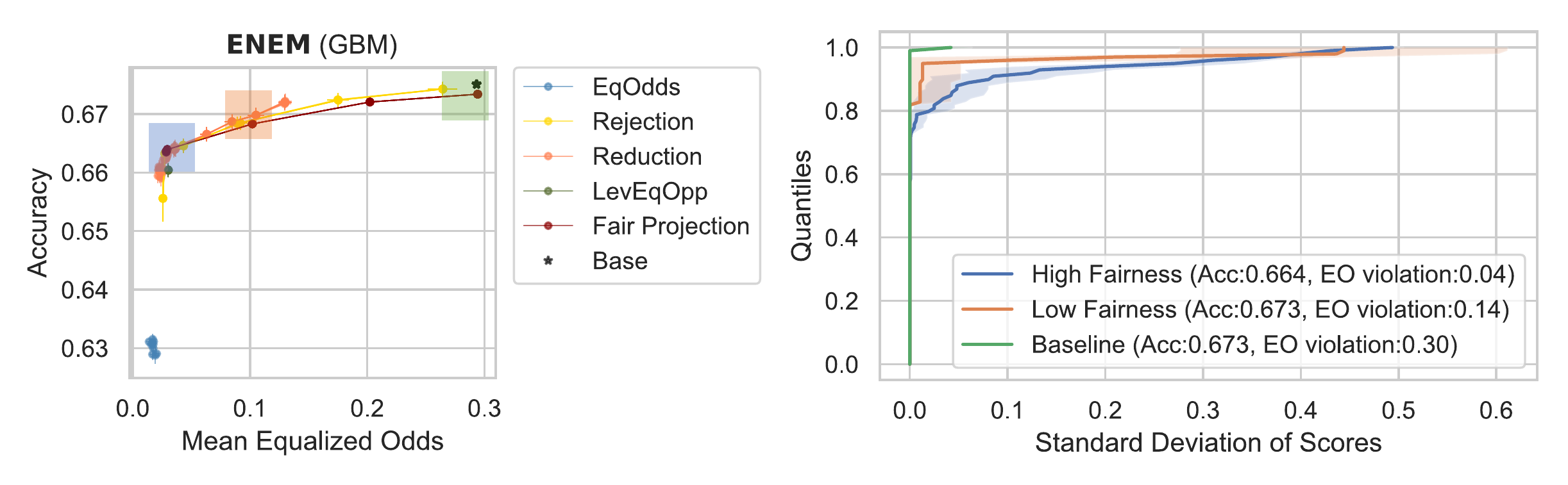}
\caption{\footnotesize \textbf{Left}: Accuracy-fairness curves of baseline gradient boosting models (GBM) v.s. fair models on the ENEM dataset.
\textbf{Right}: Quantiles of per-sample score std. across high/low fairness models and baseline.}
\end{figure}

\begin{figure}[!tb]
\centering
\includegraphics[width=1.0\textwidth]{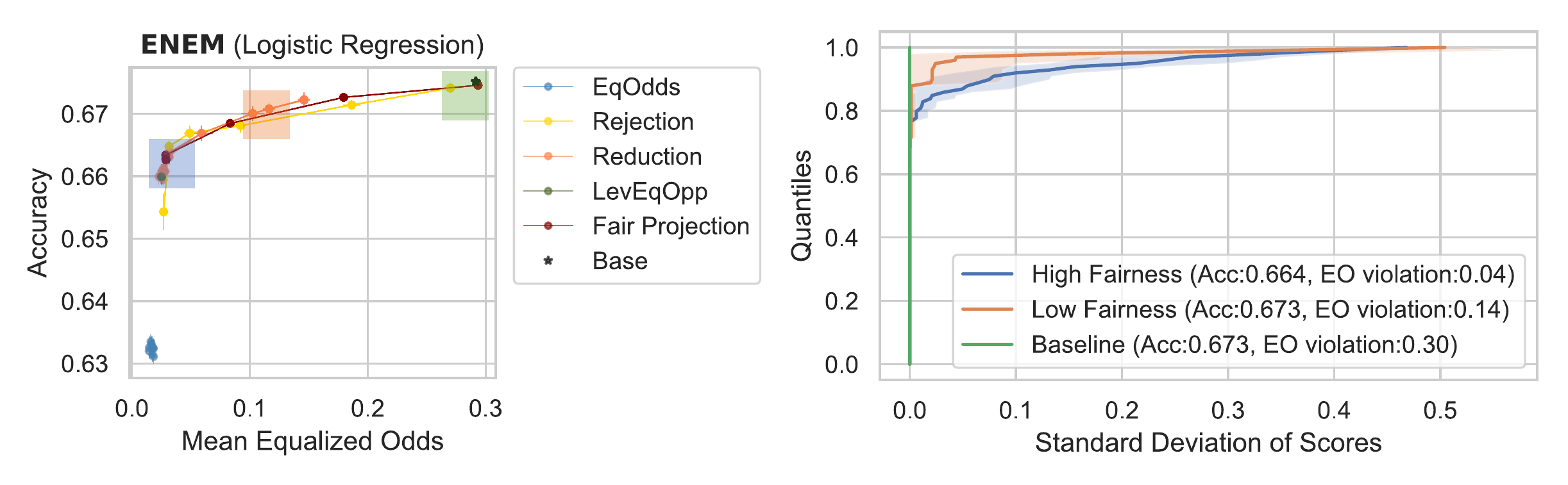}
\caption{\footnotesize \textbf{Left}: Accuracy-fairness curves of baseline logistic regression models v.s. fair models on the ENEM dataset.
\textbf{Right}: Quantiles of per-sample score std. across high/low fairness models and baseline.}
\end{figure}

\begin{figure}[!tb]
\centering
\includegraphics[width=1.0\textwidth]{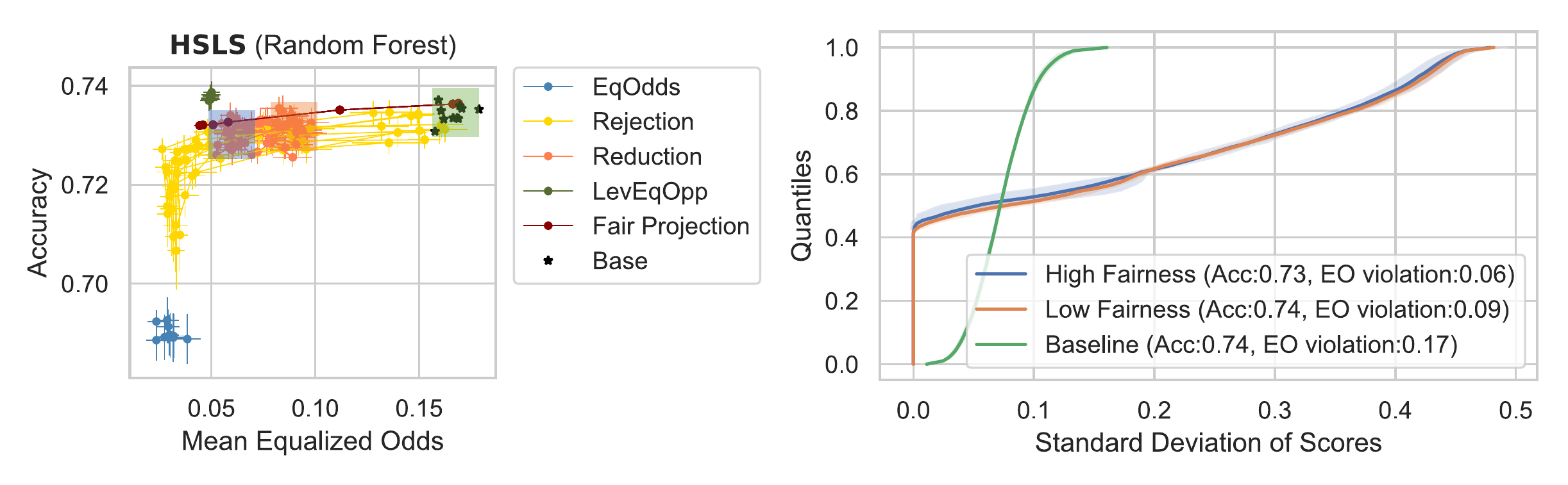}
\caption{\footnotesize \textbf{Left}: Accuracy-fairness curves of baseline random forest models v.s. fair models on the HSLS dataset.
\textbf{Right}: Quantiles of per-sample score std. across high/low fairness models and baseline.}
\end{figure}

\begin{figure}[!tb]
\centering
\includegraphics[width=1.0\textwidth]{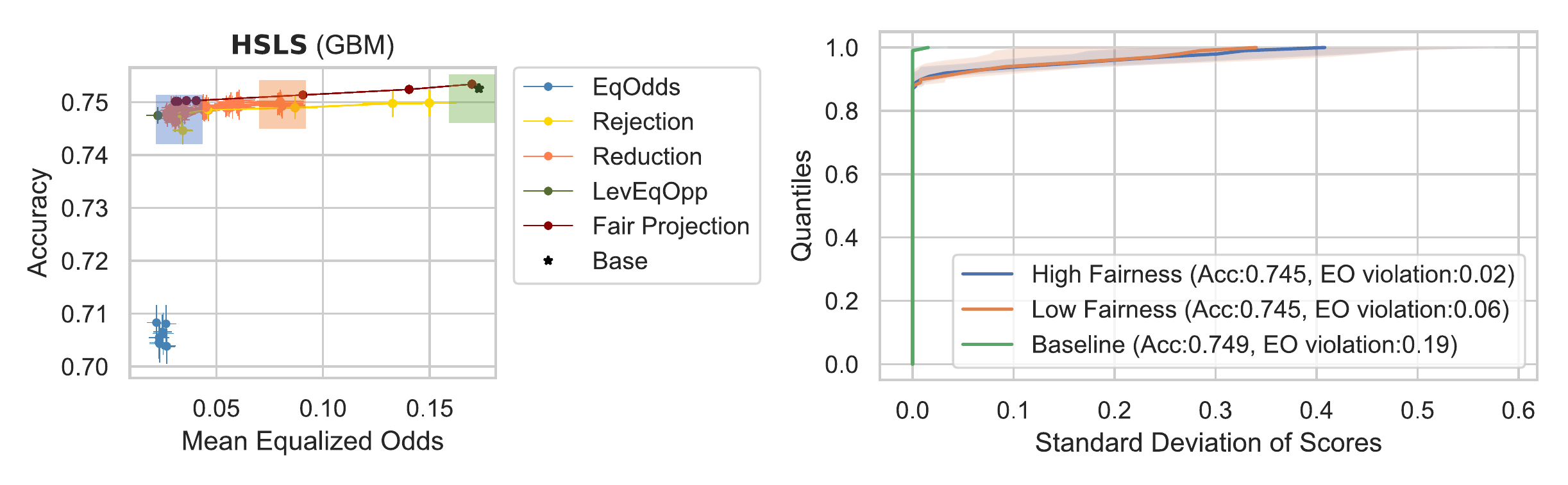}
\caption{\footnotesize \textbf{Left}: Accuracy-fairness curves of baseline gradient boosting models (GBM) v.s. fair models on the HSLS dataset.
\textbf{Right}: Quantiles of per-sample score std. across high/low fairness models and baseline.}
\end{figure}

\begin{figure}[!tb]
\centering
\includegraphics[width=1.0\textwidth]{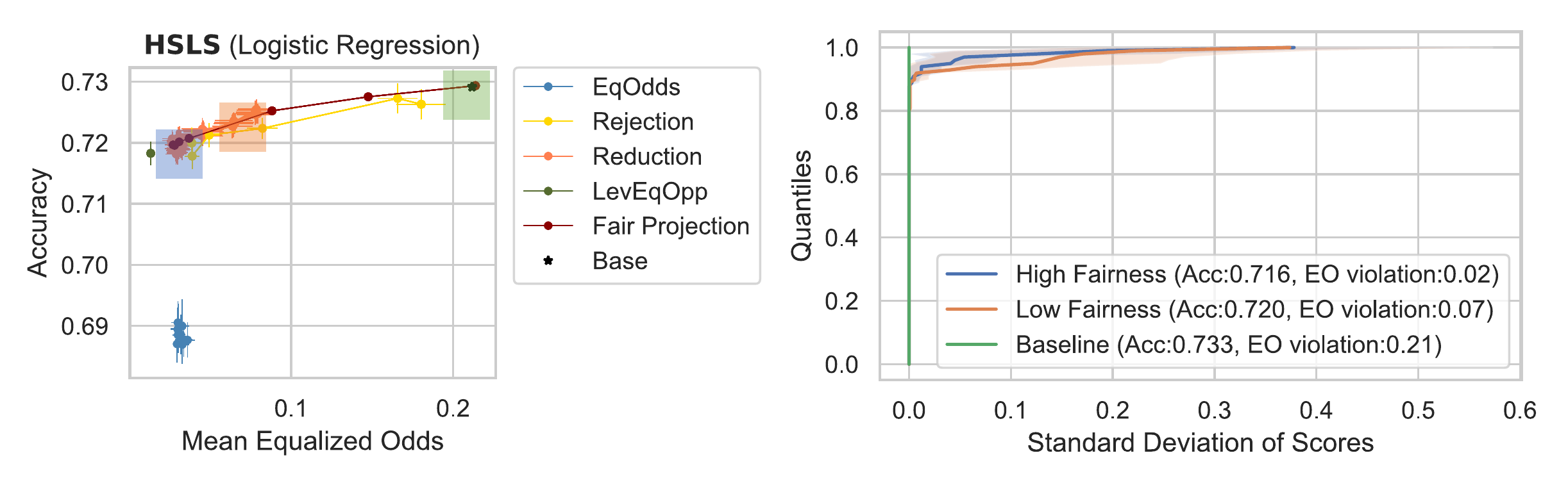}
\caption{\footnotesize \textbf{Left}: Accuracy-fairness curves of baseline logistic regression models v.s. fair models on the HSLS dataset.
\textbf{Right}: Quantiles of per-sample score std. across high/low fairness models and baseline.}
\end{figure}

\begin{figure}[!tb]
\centering
\includegraphics[width=1.0\textwidth]{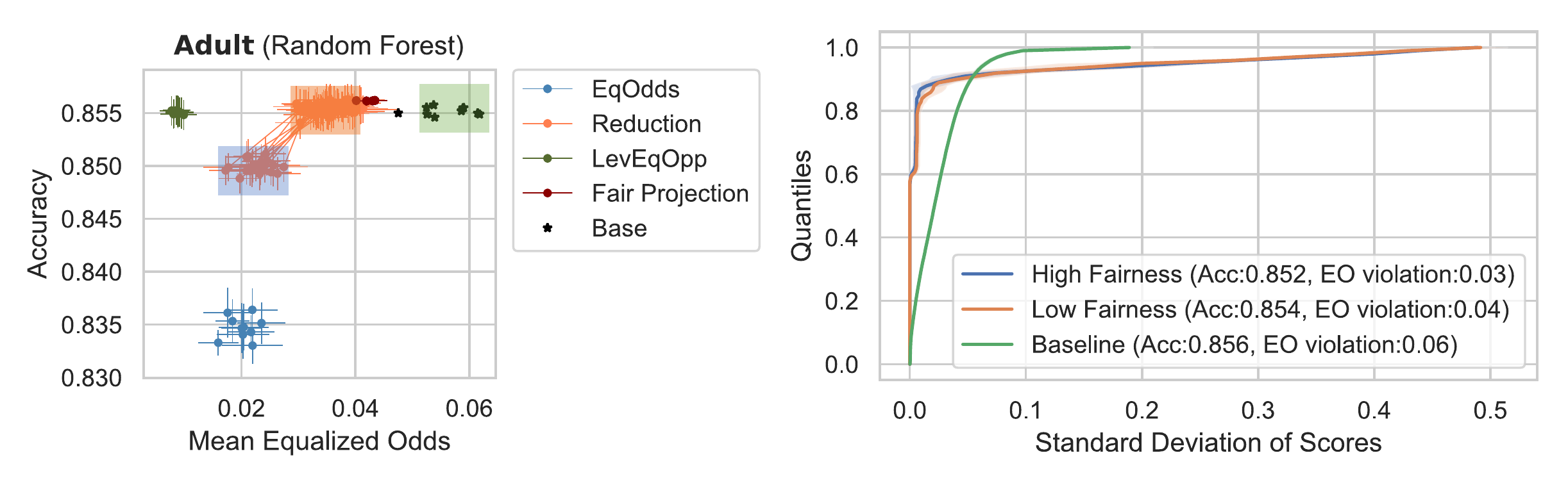}
\caption{\footnotesize \textbf{Left}: Accuracy-fairness curves of baseline random forest models v.s. fair models on the Adult dataset.
\textbf{Right}: Quantiles of per-sample score std. across high/low fairness models and baseline.}
\end{figure}

\begin{figure}[!tb]
\centering
\includegraphics[width=1.0\textwidth]{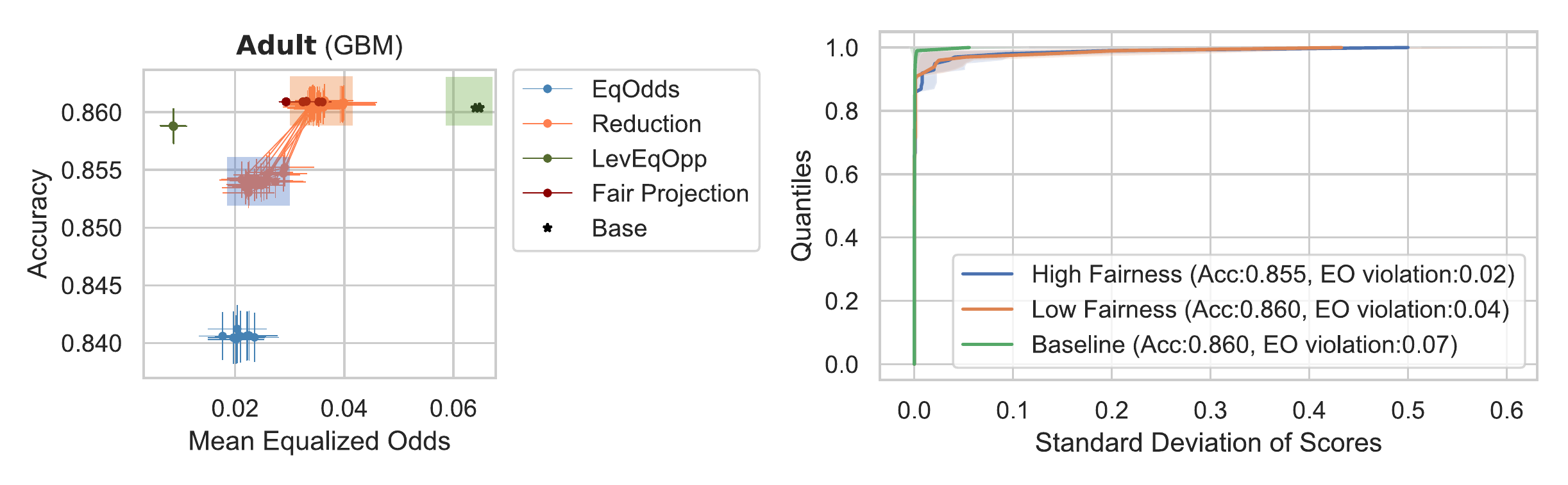}
\caption{\footnotesize \textbf{Left}: Accuracy-fairness curves of baseline gradient boosting models (GBM) v.s. fair models on the Adult dataset.
\textbf{Right}: Quantiles of per-sample score std. across high/low fairness models and baseline.}
\end{figure}

Furthermore, we evaluate the predictive multiplicity of models corrected by different fairness intervention methods. Across datasets, all fairness intervention methods exhibit maximal standard deviation of scores of 0.5 at top quantiles for random forest baseline methods. \textsc{Leveraging} \citep{chzhen2019leveraging} exhibit score arbitrariness comparable to that of baseline for GBM and Logistic Regression methods. \textsc{Rejection} and \textsc{Leveraging} output thresholded scores directly, while \textsc{Reduction} outputs probabilities (with most scores close to 0 or 1). 

\begin{figure}[!tb]
\centering
\includegraphics[width=1.0\textwidth]{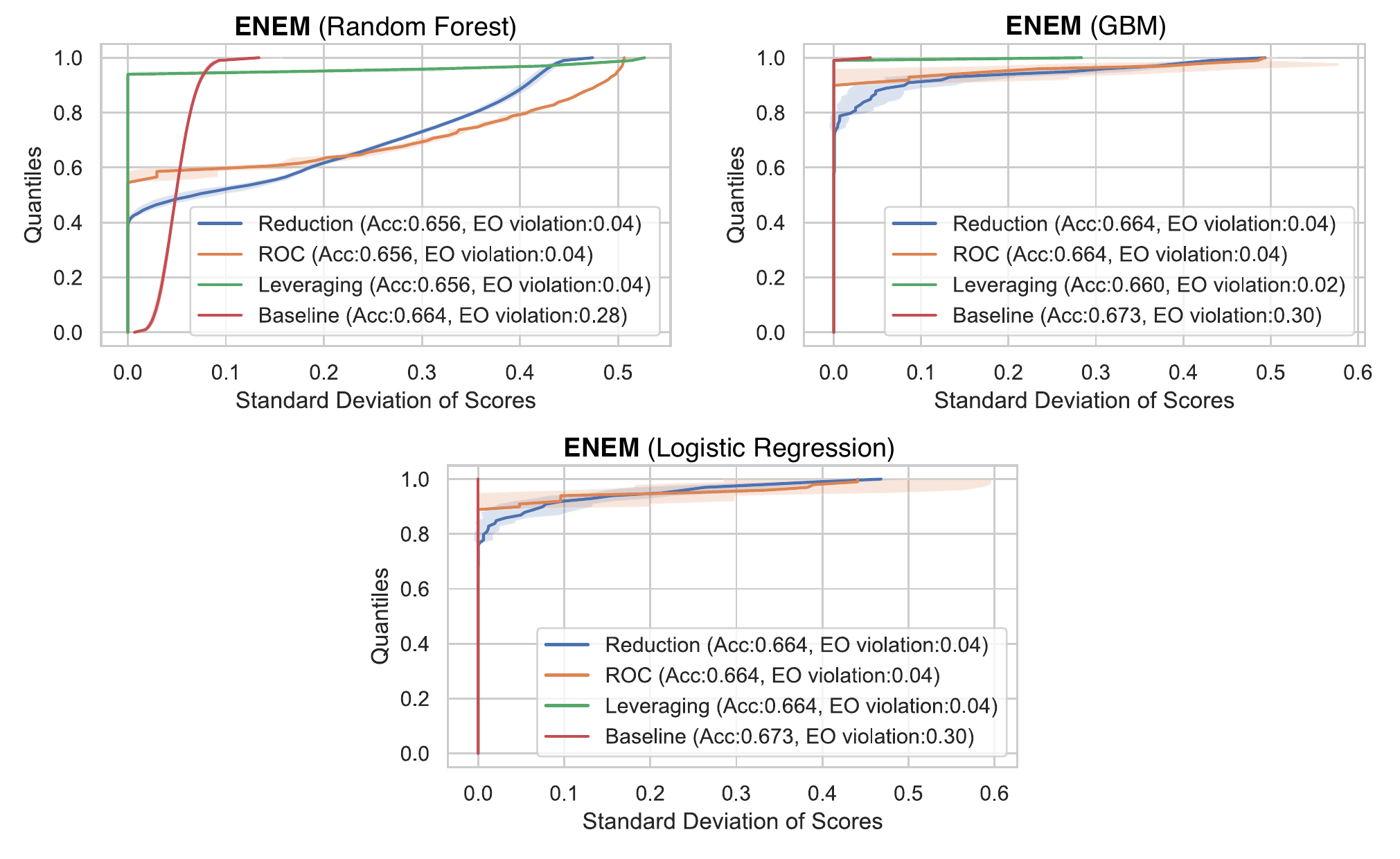}
\caption{\footnotesize
Quantile plot on models in high-fairness bin for various fairness interventions v.s. baseline models on ENEM. Fair models produce larger score std. at top percentiles compared to the baseline model.}
\end{figure}

\begin{figure}[!tb]
\centering
\includegraphics[width=1.0\textwidth]{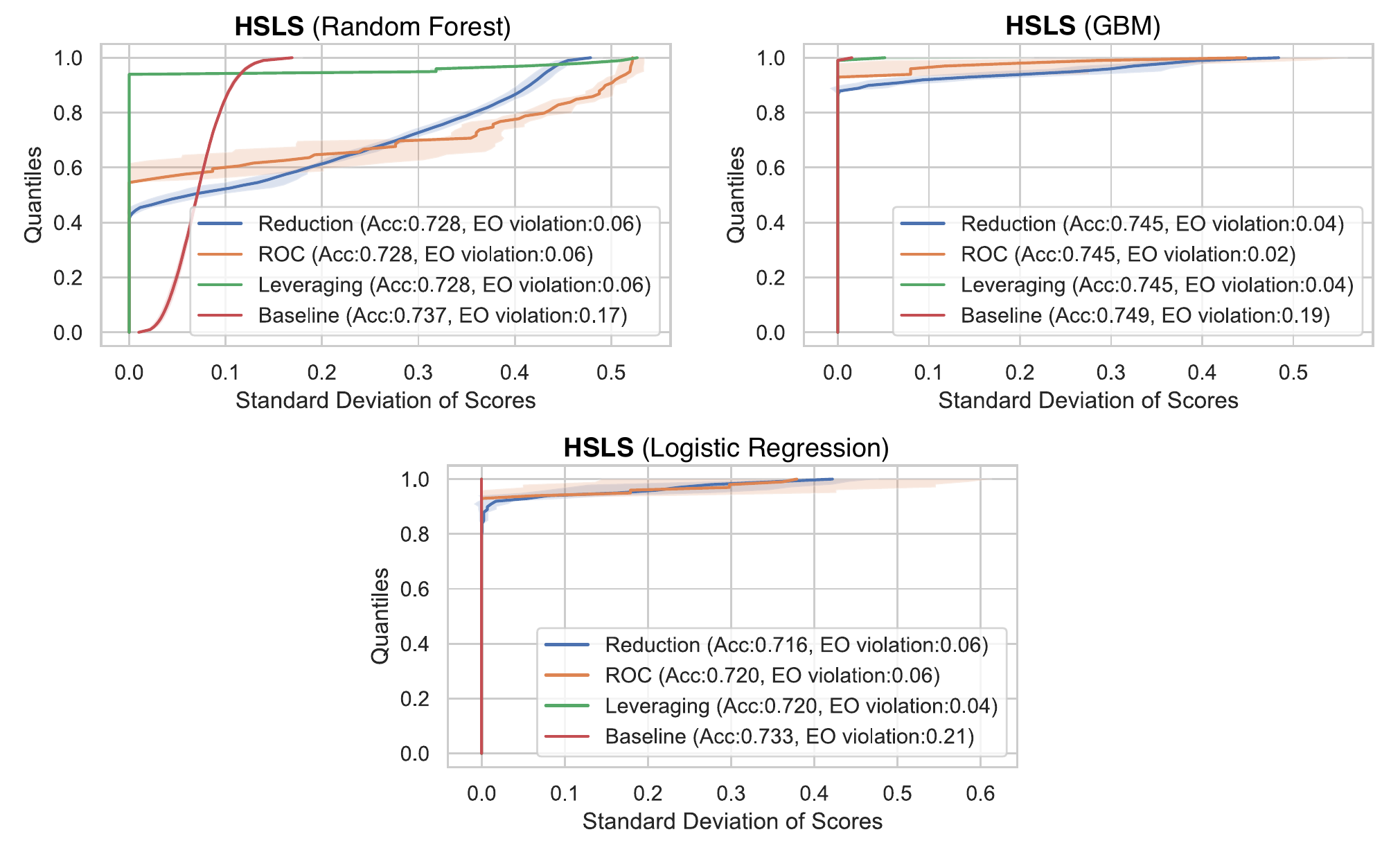}
\caption{\footnotesize
Quantile plot on models in high-fairness bin for various fairness interventions v.s. baseline models on HSLS. Fair models produce larger score std. at top percentiles compared to the baseline model.  }
\end{figure}



\end{document}